\documentclass[lettersize,journal, ]{IEEEtran}
\usepackage{amsmath,amsfonts}
\usepackage{algorithmic}
\usepackage{array}
\usepackage{textcomp}
\usepackage{stfloats}
\usepackage{url}
\usepackage{verbatim}
\usepackage{graphicx}
\def\BibTeX{{\rm B\kern-.05em{\sc i\kern-.025em b}\kern-.08em
    T\kern-.1667em\lower.7ex\hbox{E}\kern-.125emX}}
\usepackage{balance}
\usepackage{verbatim}
\usepackage{amsfonts}
\usepackage{amssymb}
\usepackage{stfloats}
\usepackage{cite}
\usepackage{soul}
\usepackage{setspace}
\usepackage{graphicx}
\usepackage{psfrag}
\usepackage{amsmath}
\usepackage{array}
\usepackage{epstopdf}
\usepackage{authblk}
\usepackage{graphicx} 
\usepackage{amsthm} 
\usepackage{lipsum}
\usepackage{verbatim} 
\usepackage{authblk}
\usepackage{mathtools}
\usepackage{cuted}
\usepackage[lined,boxed,ruled]{algorithm2e}
\usepackage{booktabs}
\usepackage{subfigure}

\usepackage{hyperref}
\hypersetup{colorlinks=false,
            linkcolor=red,
            anchorcolor=green,
            citecolor=black}

\usepackage[usenames]{color}

\newcommand{\blue}{\textcolor{blue}}

\newtheorem{Definition}{Definition}

\newtheorem{Lemma}{Lemma}
\newtheorem{Theorem}{Theorem}

\newtheorem{Assumption}{Assumption}

\newtheorem{Remark}{Remark}

\begin{document}

\title{Spectrum Breathing: Protecting Over-the-Air Federated Learning Against Interference}

\author{{Zhanwei~Wang}, {Kaibin~Huang}, and {Yonina~C.~Eldar}

\thanks{Z. Wang and K. Huang are with the Department of Electrical and Electronic Engineering, The University of Hong Kong (HKU), Hong Kong (Email: \{zhanweiw, huangkb\}@eee.hku.hk). Y. C. Eldar is with the Faculty of Math and CS, Weizmann Institute of Science, Rehovot, Israel (Email: yonina.eldar@weizmann.ac.il). The corresponding author is K. Huang.}}



\maketitle

\begin{abstract}
\emph{Federated Learning} (FL) is a widely embraced paradigm for distilling artificial intelligence from distributed mobile data.
However, the deployment of FL in mobile networks can be compromised by exposure to interference from neighboring cells or jammers.
Existing interference mitigation techniques require multi-cell cooperation or at least interference channel state information, which is expensive in practice. 
On the other hand, power control that treats interference as noise may not be effective due to limited power budgets, and also that this mechanism can trigger countermeasures  by  interference sources.
As a practical approach for protecting FL against interference, we propose \emph{Spectrum Breathing}, which cascades stochastic-gradient pruning and spread spectrum to suppress interference without bandwidth expansion.
The cost is higher learning latency by exploiting the graceful degradation of learning speed due to pruning.
We synchronize the two operations such that their levels are controlled by the same parameter, \emph{Breathing Depth}.
To optimally control  the parameter, we develop a martingale-based approach to convergence analysis of Over-the-Air FL with spectrum breathing, termed  AirBreathing FL.
We show a performance tradeoff between gradient-pruning and interference-induced error as regulated by the breathing depth.
Given receive SIR and model size, the optimization of the tradeoff yields two schemes for controlling the breathing depth that can be either fixed or adaptive to channels and the learning process.
As shown by experiments, in scenarios where traditional Over-the-Air FL fails to converge in the presence of strong interference, AirBreahing FL with either fixed or adaptive breathing depth can ensure convergence where the adaptive scheme achieves close-to-ideal performance.
\end{abstract}


\section{Introduction}

A key operation of the \emph{sixth-generation} (6G) mobile network is to distill intelligence from enormous mobile data at the network edge using distributed machine learning algorithms, resulting in an active area termed edge learning \cite{RN245,RN171}.
The obtained \emph{Artificial Intelligence} (AI) is expected to empower many \emph{Internet-of-Things} (IoT) applications ranging from smart cities to auto-pilots to extended reality.
\emph{Federated Learning} (FL) is arguably the most popular edge-learning framework for its preservation of data ownership and being considered for the 6G standard \cite{RN259,gafni2022federated}.
FL protects users' data privacy by distributing the learning task and requiring users to upload local model updates instead of raw data \cite{RN200}.
Among others, two key challenges stymieing the deployment of FL in a mobile network are 
1) a communication bottleneck resulting from the transmission of high-dimensional model updates and 
2) exposure to interference from neighboring cells, and jammers \cite{RN171,RN143,RN211,RN226}.
To simultaneously tackle these two challenges, we propose a spectrum-efficient method for suppressing interference to FL in mobile networks, termed \emph{Spectrum Breathing}. 

The key operation of FL is for a server to upload local updates from devices, which are computed using local data, for aggregation to update the global model.
To overcome the resultant communication bottleneck, previous works focus on designing task-oriented wireless techniques for FL with the aim to alleviate the effects of channel hostility on learning performance.
Diversified approaches have been proposed including  radio resource management \cite{RN24,RN216}, power control \cite{RN125,RN10}, and device scheduling \cite{RN247,RN160}.
On the other hand, the direct approach to reduce communication overhead is to prune local model updates, namely local models or stochastic gradients, and furthermore adapt the pruning operation to wireless channels \cite{RN149,RN231, RN193}.
Instead of incurring unrecoverable distortion, gradient pruning can be translated via randomization into longer learning latency with small accuracy degradation \cite{RN229}.

Recently, a new class of techniques, termed \emph{Over-the-Air FL} (AirFL), has emerged to address the scalability issue in multi-access  by many devices under a constraint on radio resources \cite{RN34,RN234,RN11}.
Underpinning AirFL is the use of so-called \emph{Over-the-Air Computing} (AirComp) to realize over-the-air aggregation of local updates by exploiting the waveform superposition property of a multi-access channel, and thereby enable simultaneous access \cite{guangxu-WC}.
Building on AirComp, the efficiency of AirFL can be enhanced by beamforming \cite{RN234}, gradient pruning \cite{RN11}, broadband transmission \cite{RN34} and power control \cite{RN10,RN122}, and even the use of \emph{Intelligent Reflecting Surface} (IRS) \cite{RN250}.
Different from traditional designs, such techniques aim to realize the required signal-magnitude alignment at the server to implement AirComp despite channel distortion.
For instance, an IRS can help to overcome such distortion to suppress the  alignment error \cite{RN250}.
Furthermore, the optimization of AirFL techniques enables them to be adapted to not only channel states but also  learning operations (e.g., gradient statistics in the current round \cite{RN10}).
The effectiveness of AirFL and AirComp at large hinges on the use of uncoded linear analog modulation for transmission.
This exposes AirFL to interference and gives rise to the challenge of how to make AirFL robust.
Gradient pruning techniques mentioned earlier do not address this challenge
as they merely reduce communication overhead without making any attempt at interference suppression.
An alternative approach is to treat interference as noise and regulate it using existing power-control techniques for AirFL \cite{RN133,RN309}.
This class of techniques' effectiveness in dealing with interference is limited for two reasons.
First, interference power may be comparable with that of the signal if not larger and far exceeds the noise power.
Second, unlike noise, an interference source (e.g., a neighboring access point or a jammer) is active and can react to the power control of a signal source in a way that renders it ineffective.

An additional line of work is to adapt the rich set of existing interference-mitigation techniques to suit AirFL \cite{RN143,RN211,RN251,RN159,RN252}.
Previous works share the common principle of relying on cooperation between interfering nodes to mitigate the effects of their mutual interference on the learning performance.
This principle is materialized in diversified techniques for multi-cell AirFL systems, including spatial interference cancellation \cite{RN143}, signal-and-interference alignment into orthogonal signal sub-spaces \cite{RN211},  and cooperative power control and devices scheduling \cite{RN159,RN252}.
However, their implementation requires accurate \emph{Channel State Informantion} (CSI) of interference channels.
Acquiring such information can incur extra overhead and latency due to inter-cell messaging in multi-cell systems, and is infeasible in scenarios where the interference sources are in other networks or jammers.

One classic interference mitigation technique, called spread spectrum, has no such limitations but has not been explored in the context of AirFL due to its low efficiency in spectrum utilization\cite{RN20}.
This technique can reduce interference power by a factor, called the \emph{Processing Gain} denoted as $G$, if a narrowband signal is spread in the spectrum by $G$ via scrambling using a \emph{Pseudo-Noise} (PN) sequence at the transmitter and using the same sequence to reverse the operation, called despreading, at the receiver.
These operations neither require multi-cell cooperation nor interference CSI.
Its invention served the purpose of anti-jamming for secured communication in World War II while its commercial success was due to the use for mitigating multi-user interference in a resultant multi-access scheme, termed \emph{Code-Divison Multi-Access} (CDMA), for 3G \cite{RN20,RN36}.

Gradient pruning and spread spectrum are two well-known techniques. The novelty of the proposed spectrum breathing approach lies in their integration to cope with interference in an AirFL system under a bandwidth constraint.
Specifically, deployed at each device, the technique cascades two operations before transmission -- \emph{random pruning} of local gradient, called spectrum contraction, and \emph{spread spectrum} on the pruned gradient.
    Note that random pruning is more suitable for AirFL than the alternative of magnitude-based pruning \cite{RN192} (also see discussion in Sec. \ref{Sec: Distortion}).
    The spectrum contraction and spreading are governed by parameter, call \emph{breathing depth}.
    As mentioned, the former mainly results in lengthened learning latency; the latter suppresses interference power by the factor of breathing depth.
    As a result, AirFL can converge even in the presence of strong interference.
    In the iterative FL algorithm, the alternating spectrum contraction and spreading are analogous to human breathing, giving the technique its name.
We optimally control the spectrum breathing parameter so as to maximize its performance gain.

The contributions of this paper are summarized as follows.

\begin{itemize}
    \item \textbf{Convergence Analysis:} Adjusting the breathing depth provides a mechanism for controlling AirFL using spectrum breathing, termed AirBreathing FL.
    To facilitate optimal control, we analyze the learning convergence by extending an existing supermartingale-based approach to account for AirComp, spectrum breathing and fading.
    The derived results reveal a tradeoff as regulated by the breathing depth.
    Specifically, increasing the parameter has two opposite effects -- one is to improve the successful convergence probability by interference suppression and the other is to decrease it due to more aggressive gradient pruning.
    This gives rise to the need of optimal control.
    \item \textbf{Control of Spectrum Breathing:} Using the preceding tradeoff, the optimization of breathing depth yields two schemes for controlling AirBreathing FL under given receive SIR and model size.
    First, without CSI and \emph{Gradient State Information} (GSI) at the server, the parameter is fixed over rounds and its optimal value is derived in closed form. 
    Second,  when CSI and GSI are available as in \cite{RN10}, the optimal strategy is designed to be adapted to CSI and GSI.
    \item \textbf{Experimental Results:} The results from experiments on AirBreathing FL demonstrate satisfactory learning performance even in the cases with strong interference that could fail the learning task without  spectrum breathing. 
    Moreover, spectrum breathing with depth adaptation outperforms the case with a fixed breathing depth.    
\end{itemize}

The remainder of this paper is organized as follows. Models and metrics are introduced in Sec. \ref{Sec:ModelandMetrics}. 
The effects of pruning on generic data and FL is demonstrated in Sec. \ref{Sec: Distortion}.
Overview design of spectrum breathing is illustrated in Sec. \ref{Sec:SBsys}.
Convergence analysis and breathing depth optimization are analyzed in Sec. \ref{Sec:ConvergenceAnalysis} and \ref{Sec:optmization}, respectively. 
Experimental results are provided in Sec. \ref{Sec:experiments}, followed by concluding remarks in Sec. \ref{Sec:conclusion}.

%

\section{Models and Metrics}\label{Sec:ModelandMetrics}

\begin{figure*}[t]
	\centering
	\begin{minipage}[b]{0.7\textwidth}
		\centering
		\includegraphics[width=\textwidth]{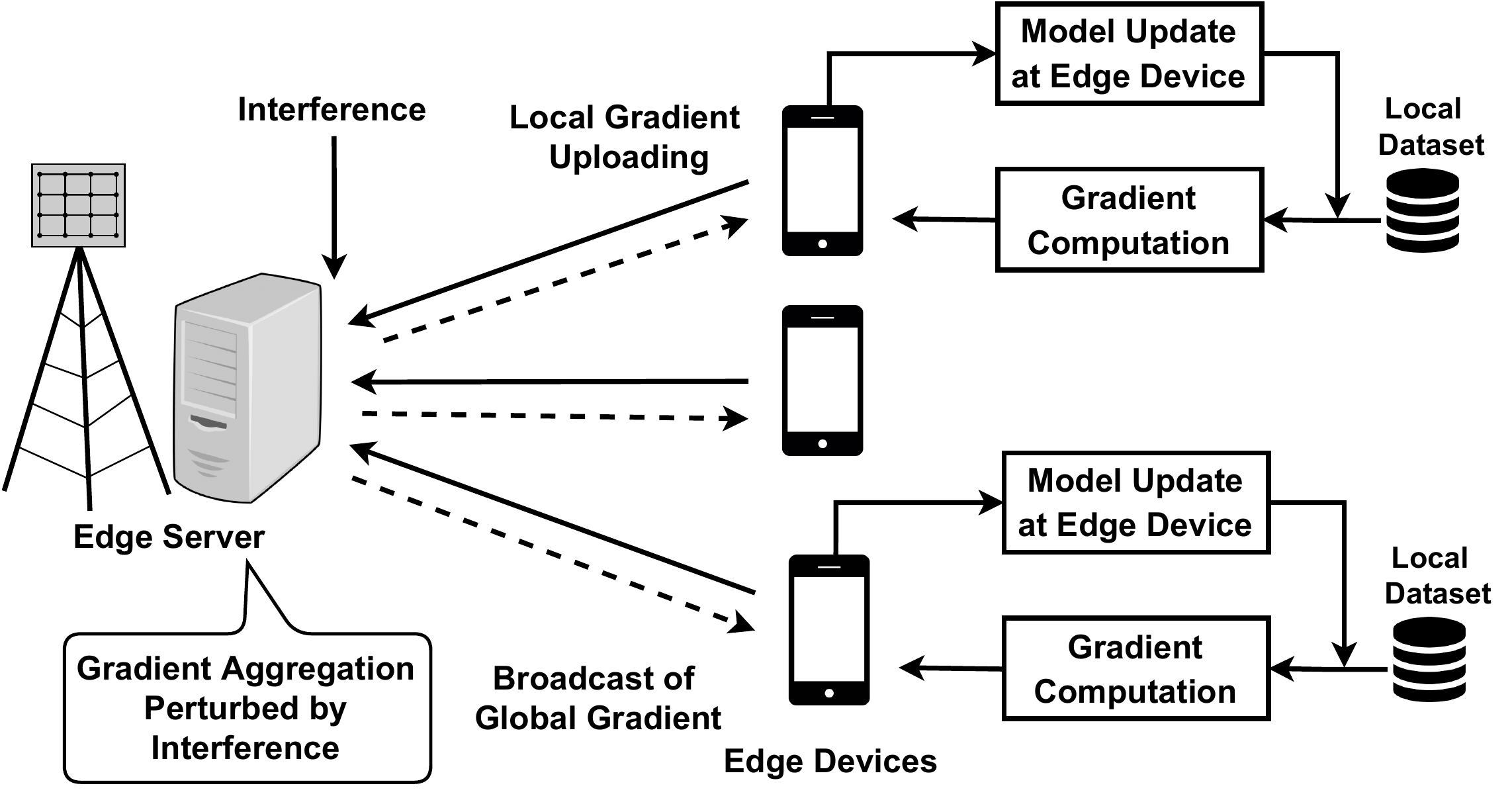}
	\end{minipage}
	\caption{System diagram of AirFL system perturbed by interference.} 
	\label{fig:system-diagram}

\end{figure*}

We consider an AirFL system as illustrated in Fig. \ref{fig:system-diagram} that comprises one server (collocated with base station) and $K$ devices.
The learning process is perturbed by external interference (e.g. from other cells). The learning and communication models are described separately in the following sub-sections.

\subsection{Learning Model}

We first describe the FL process underpinning AirFL. Each device, say $k$, maintains its
 local dataset $\mathcal{D}_k$ including $|\mathcal{D}_k|$ pairs of data sample $\mathbf{x}_j$ and label $y_j$, denoted as $\{(\mathbf{x}_j,y_j)\}\in \mathcal{D}_k,j\in\{1,2,\dots,|\mathcal{D}_k|\}$. 
The server coordinates $K$ devices to optimize the weights of the global model $\mathbf{w}\in \mathbb{R}^{D}$ where $D$ is the model size, under the criterion of minimizing a global loss defined as
\begin{equation}
\label{loss function}
    F(\mathbf{w})\triangleq \frac{1}{\sum_{k=1}^K|\mathcal{D}_k|}\sum_{k=1}^K\sum_{(\mathbf{x}_j,y_j)\in \mathcal{D}_k} f(\mathbf{w},\mathbf{x}_j,y_j),
\end{equation}
where $f(\mathbf{w},\mathbf{x}_j,y_j)$ is the empirical loss function  indicating the prediction error on model $\mathbf{w}$ using a data sample $(\mathbf{x}_j,y_j)$.
For simplicity, we denote $f(\mathbf{w},\mathbf{x}_j,y_j)$ as $f_j(\mathbf{w})$. 
Distributed \emph{Stochastic Gradient Descent} (SGD) is applied to minimize the global loss. Specifically, time is divided into $N$ rounds with index $n\in \{0,1,\dots,N-1\}$.
Considering round $n$, each device computes the gradient of the empirical loss function using a mini-batch of local dataset. The gradient of device $k$  is given as
\begin{equation}
    \mathbf{g}_k(n)= \frac{1}{|\mathcal{B}_k|}\sum_{j\in \mathcal{B}_k}\nabla f_j(\mathbf{w}(n)),
    \label{grad}
\end{equation}
where $\mathcal{B}_k \subseteq \mathcal{D}_k$ is the selected mini-batch of $\mathcal{D}_k$, and $\nabla$ represents the gradient operation. If local gradients can be reliably transmitted to the server, the global estimate of the gradient of the loss function in (\ref{loss function}) is obtained as
\begin{equation}
    \overline{\mathbf{g}}(n)=\frac{1}{K}\sum_{k=1}^K\mathbf{g}_k(n).
\end{equation}
Then, $\overline{\mathbf{g}}(n)$ is broadcast back to each device, by which the current model is updated via gradient descent:
\begin{equation}
    \mathbf{w}(n+1)=\mathbf{w}(n)-\eta\cdot \overline{\mathbf{g}}(n),
    \label{gsgd}
\end{equation}
where $\eta$ denotes the learning rate. The distributed SGD is thus to iterate (\ref{grad})
 and (\ref{gsgd}) until a convergence condition is met.


 \subsection{Communication Model}

The uploading of the gradients using AirComp is perturbed by interference. To combat interference, each transmitted signal undergoes the operations of random pruning and spread spectrum. The two operations of AirBreathing FL are elaborated in Section \ref{Sec:SBsys}. For the current exposition, some useful notation is introduced. Considering round $n$, the $s$-th element of the pruned local gradient transmitted by device $k$ is scrambled by spread spectrum into a sequence, denoted $\widetilde{\mathbf{g}}_{k,s}(n)$, with each element called a chip and the $\ell$-th chip denoted as $[\widetilde{\mathbf{g}}_{k,s}(n)]_\ell$.

Using the above notation, AirComp can be modelled as follows. Assume  chip-level synchronization between devices using a standard technique such as 
Timing Advance \cite{TA_1}. The simultaneous transmission of the $(s,\ell)$-th chips, i.e., $[\widetilde{\mathbf{g}}_{k,s}(n)]_\ell$, enables AirComp to yield the corresponding received chip symbol, given as
\begin{equation}
    [\mathbf{y}_s(n)]_{\ell}=\sum_{k=1}^K h_{k}(n)p_k(n) [\widetilde{\mathbf{g}}_{k,s}(n)]_{\ell}+ [\mathbf{z}_s(n)]_{\ell}, \quad \forall 
    (s,\ell),
    \label{receivedchip}
\end{equation}
where $h_{k}(n)\sim \mathcal{CN}(0,1)$ represents the $k$-th Rayleigh fading channel coefficient that remains constant with round $n$, $p_k(n)$ the transmission power, 
$[\mathbf{z}_s(n)]_{\ell}\sim \mathcal{CN}(0,2P_I)$ is the additive complex Gaussian interference. 
We consider the worst-case interference distribution that is Gaussian over chip duration \cite{Worst-Gaussian}. Given an interference-limited system, channel noise is assumed negligible.

The downloading of the aggregated gradient can be implemented using digital or analog transmission \cite{RN106}. Besides the availability of full bandwidth, transmission power at the server is much larger than that of the devices. Thus, gradient broadcasting is much more reliable than local gradient uploading such that the distortion to downlink is negligible.

\subsection{Performance Metric}
To quantify the distortion from gradient-pruning and interference, we introduce an AirComp error. Consider round $n$ and active device set $\mathcal{K}(n) \subseteq\{1,2, \ldots, K\}$.
After post-processing the received signal $\mathbf{y}(n)$ in (\ref{receivedchip}), the output at the server is denoted as $\mathbf{y}^\prime(n)$ specified in Sec. \ref{Sec:SBsys}. 
Then AirComp error is defined as the \emph{Mean Squared Error} (MSE) between $\mathbf{y}^\prime(n)$ and its desired ground-truth, namely $\frac{1}{|\mathcal{K}(n)|} \sum_{k \in \mathcal{K}(n)} \mathbf{g}_k(n)$, as
\begin{equation}
\label{MSE_metric}
\begin{split}
    \textup{MSE}(n)&=\mathbb{E}\left[   \left\Vert \frac{1}{|\mathcal{K}(n)|}\sum_{k\in\mathcal{K}(n)}\mathbf{g}_k(n)-\mathbf{y}^\prime(n) \right\Vert^2 \right],\\
\end{split}
\end{equation}
where the expectation is taken over the distributions of the transmitted symbols, interference, channel fading, and pruning pattern.


\section{Distortion from Pruning - Generic Data versus Federated Learning}
\label{Sec: Distortion}

The spectrum-contraction operation of AirBreathing FL is realized via pruning local gradients in the FL process. Its effect on the system performance is fundamentally different from that of pruning a generic data sequence. This can be better understood by analyzing and comparing the two effects in the remainder of this section.

\subsection{Generic Data Pruning}
Consider gradient $\mathbf{g}_k(n) \in \mathbb{R}^D $, that is i.i.d. distributed Gaussian vector with each element having zero-mean and variance of $\sigma^2$. The desired ground truth $ \frac{1}{|\mathcal{K}(n)|}\sum_{k\in \mathcal{K}(n) }\mathbf{g}_k(n)$ is compressed by random pruning, namely a function that randomly  replaces elements with zeros. Let $\mathbf{g}^{\text{sp}}(n)$ and $\gamma$ denote the pruned gradient and the remaining fraction of nonzero elements, called the pruning ratio.
For generic data, the distortion from pruning is commonly quantified as the MSE between the pruned sequence and its ideal version:
\begin{equation}
\begin{split}
    \text{MSE}(n)&=\mathbb{E}\left[\left\Vert \mathbf{g}^{\text{sp}}(n)- \frac{1}{|\mathcal{K}(n)|}\sum_{k\in \mathcal{K}(n) }\mathbf{g}_k(n) \right\Vert^2 \right]\\
    &=(1-\gamma) \frac{D\sigma^2}{|\mathcal{K}(n)|}  .
\end{split}
\end{equation}
One can see that the distortion increases linearly with the level  of pruning $(1-\gamma)$.
If the pruning represents channel erasures, then the lost information can not be recovered at the server.

\subsection{Stochastic Gradient Pruning}

The reliability of a generic communication system is measured by data distortion as we have discussed.
On the contrary, the performance of an FL system is measured using an \emph{End-to-End} (E2E) metric such as convergence rate or learning accuracy.
In such a system, the pruning of transmitted data has the effect of slowing down the learning speed.
Recall that FL is essentially a distributed implementation of SGD.
To substantiate the above claim, we consider randomly pruned SGD implemented using classic \emph{Block Coordinate Descent} (BCD) \cite{RN229}.
Let the local loss function, $f_j(\mathbf{w})$, comprise a smooth and convex loss function, $\hat{f}_j(\mathbf{w})$, that is regularized by a block separable function, $\Phi_j(\mathbf{w})$:
\begin{equation}
  f_j(\mathbf{w}) =\hat{f}_j(\mathbf{w})+\Phi(\mathbf{w}).
\end{equation}
The regularization $\Phi(\mathbf{w})=\sum_{b=1}^B\Phi_{b}(\mathbf{w}_b)$ is a sum of $B$ convex, closed functions $\Phi_{b}(\mathbf{w}_b)$ corresponding to the selected block of model parameters $\mathbf{w}_b$.
The blocks are non-overlapping and together they constitute the whole model.
Considering round $n$, the server selects one block randomly and notifies devices. Then each device computes the gradient locally based on $\mathbf{w}(n)$ and   uploads the specified block of coefficients to the server. Thus, only one  selected block is updated using a pruned gradient aggregated from devices, i.e., $\mathbf{g}^{\text{sp}}(n)$, while others remain unchanged. Mathematically,
\begin{equation}
   \mathbf{w}(n+1)=\mathbf{w}(n) - \eta \mathbf{g}^{\text{sp}}(n).
\end{equation}
Equivalently,  BCD can be seen as distributed SGD updated with pruned gradients where the pruning ratio is $\gamma=\frac{1}{B}$.
Its convergence rate is measured by the required number of iterations, say $N_{\epsilon,\rho}$, guaranteeing $\epsilon$-accuracy  with probability of at least $1-\rho,\rho\in(0,1]$:
\begin{equation}
    \text{Pr}\{F(\mathbf{w}(N_{\epsilon,\rho}))-F(\mathbf{w}^*) \leq \epsilon \}\geq 1-\rho,
\end{equation}
where $\mathbf{w}^*$ represents the global optimality point.
It can be proved that \cite{RN229},
\begin{equation}
    N_{\epsilon,\rho}\leq \mathcal{O}\left( \frac{B}{\epsilon}\log \left(\frac{1}{\rho}\right) \right) {=}
    \mathcal{O}\left( \frac{1}{ \gamma \epsilon }\log \left(\frac{1}{\rho}\right) \right).
\end{equation}
One can observe that the required number of iterations (i.e, learning latency) is inversely proportional to the pruning ratio.

\subsection{Why Random Pruning for AirFL?}
\label{RandRrune for AirBreathing}
Random gradient/model pruning is popularly adopted for FL (see e.g., \cite{RN192, RN149}).
The alternative scheme, importance-aware pruning that prunes gradient coefficients with the smallest magnitudes, does not allow efficient implementation for several reasons discussed in \cite{RN192}.
First, AirComp requires local gradient coefficients pruned by different devices to  have identical positions in the local gradient vectors.
This cannot be guaranteed if devices perform independent importance-aware pruning.
Second, doing so requires devices to upload indices of pruned/remaining coefficients to the server to facilitate aggregation, thereby incurring additional, significant communication overhead \cite{RN270}.
Finally, importance-aware pruning increases devices' computation loads due to the coefficient sorting of high-dimensional gradient vectors.

\section{Overview of Spectrum Breathing}
\label{Sec:SBsys}

As illustrated in Fig. \ref{fig:SBtransceiver}, the proposed spectrum breathing technique consists of operations at the transmitter of a device and at the receiver of the server. They are described separately in the following sub-sections.

\subsection{Transmitter Design}
\begin{figure*}[t]
	\centering
	\begin{minipage}[b]{0.9\textwidth}
		\centering
		\includegraphics[width=\textwidth]{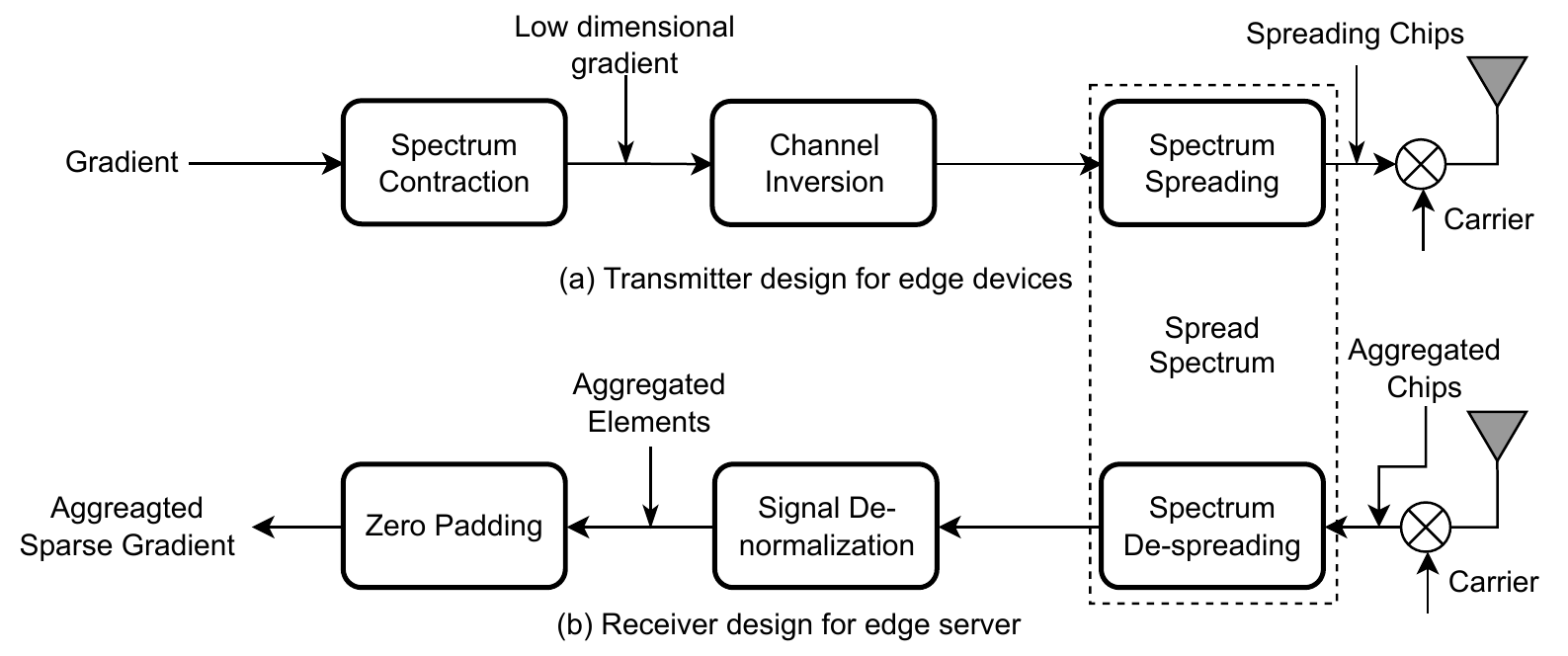}
	
	\end{minipage}
	\caption{Transceiver of the spectrum breathing system. } 
	\label{fig:SBtransceiver}

\end{figure*}

The transmitter design is shown in Fig. \ref{fig:SBtransceiver}(a), comprising three cascaded operations, i.e., spectrum contraction, channel inversion, and spectrum spreading. 
\subsubsection{Spectrum Contraction}
The operation is to randomly prune the elements of a local gradient at each device.
The purpose is to create extra bandwidth for the latter operation of spread spectrum.
The operation compresses the spectrum required for transmitting a local gradient, giving the name of spectrum contraction.
Consider round $n$ and local gradient $\mathbf{g}_k(n)$ at device $k$. 
Let $\psi_n$, $S_n\triangleq|\psi_n|$, and $\Omega_n$ denote the selected element set, number of selected elements, and the set of all $S_n$-element subsets of $\{1,2,\dots, D\}$, respectively.
At the server, $\psi_n$ is chosen randomly from $\Omega_n$ before being broadcast to devices.
Using $\psi_n$, device $k$ compresses $\mathbf{g}_k(n)$ into an $S_n\times1$ vector, denoted as $\mathbf{g}^{\text{co}}_k( n)=\Psi(\mathbf{g}_k(n))$, using the pruning function $\Psi(\cdot): \mathbb{R}^{D} \rightarrow \mathbb{R}^{S_n}$.
Note that the pruning pattern $\psi_n$ is identical for all devices as required for AirComp to realize element alignment.
Since the gradient statistics may change over iterations, normalization is needed in each round to meet the power constraint \cite{RN140}. 
The normalized pruned gradient of the compressed version is given as
 \begin{equation}
 \label{eq:normalization}
     \widehat{\mathbf{g}}^{\text{co}}_{k}(n)=\frac{\mathbf{g}^{\text{co}}_{k}(n)-M(n)\mathbf{1}}{V(n)}, 
 \end{equation}
 where $\mathbf{1}$ is an all-one vector.
 Considering i.i.d data distribution as in \cite{RN34, RN143} and random pruning, the elements of $\mathbf{g}_k^{\text{co}}(n)$ can be modeled as identically distributed random variables over $k$ with mean $M(n)$ and variance $V^2(n)$.
 This enables normalized gradient symbol power, i.e., $\frac{1}{S_n}\mathbb{E}[\Vert \widehat{\mathbf{g}}^{co}_k(n) \Vert^2]=1$.

\subsubsection{Channel Inversion}
Following \cite{RN34}, truncated channel inversion is performed to achieve amplitude alignment as required for AirComp.
We consider block fading channels such that the channel state 
 is constant in each round. 
To avoid deep fading, device $k$ is inverted only if its gain exceeds a given threshold, denoted as $G_{th}$, or otherwise device $k$ is absent in this round by setting its power as zero.  Mathematically,
 \begin{equation}
     p_k(n)=
     \begin{cases}
     \frac{\sqrt{P_0}}{h_k(n)},&|h_k(n)|^2\geq G_{th}\\
     0,&|h_k(n)|^2< G_{th},
     \end{cases}, 
     \label{Eq:TCI}
 \end{equation}
where $P_0$ is the signal-magnitude-alignment factor. 
Transmission of each device is subject to a long-term power constraint over $N$ rounds:
\begin{equation}
\label{power-constraints}
    \mathbb{E}\left[ \sum_{n=0}^{N-1} G_nS_n p_k^2(n) \right]\leq P_{\text{max}},
\end{equation}
where the expectation is taken over the randomness of channel coefficients and transmitted symbols. 
Given (\ref{receivedchip}),  $\frac{P_0}{P_I}$ determines the receive SIR of the model-update from each device \cite{RN34}.
The probability that device $k$ avoids truncation, called activation probability, is denoted by $\xi_a$ and obtained as 
 \begin{equation}
 \label{active-prob}
     \xi_a=\text{Pr}(|h_k(n)|^2\geq G_{th})=e^{-G_{th}}.
 \end{equation}
Due to random truncation, the random set of active devices of round $n$ is denoted as $\mathcal{K}(n)$, which varies over rounds. 

\subsubsection{Spectrum Spreading}

For interference suppression, spectrum spreading \cite{RN20} is performed to expand the data bandwidth, denoted as $B_s(n)$, into the whole available bandwidth, denoted as $B_c$, using PN sequences. 
Let $T_s(n)=\frac{1}{B_s(n)}$ and $T_c=\frac{1}{B_c}$ denote the duration of one gradient symbol and one chip of PN sequences, respectively.
Then a PN sequence comprises $G_n={T_s(n)}/{T_c}$ chips, when $G_n$ is called the processing gain.
The key operation of the spreader is to upsample and scramble the input elements by the corresponding PN sequences to realize spectrum expansion.
Given the $s$-th input element of the spreader, say $[\widehat{\mathbf{g}}^{\text{co}}_k(n)]_s$, the corresponding PN sequence is represented as $\mathbf{C}_s(n)\in \mathbb{R}^G$ wherein $[\mathbf{C}_s(n)]_\ell\in \{+1,-1\}, s\in \{1,2,\dots,S_n\} , \ell \in \{1,2,\dots, G_n\}$ is the $\ell$-th chip
and is generated at the server through  Bernoulli trails with the probability 0.5. The set of PN sequences, denoted as $\mathcal{C}(n)\triangleq\{\mathbf{C}_1(n),\dots,\mathbf{C}_{S_n}(n)\}$, is broadcast to all devices.
For device $k$, the output of the spreader is represented by a $G_nS_n\times1$ vector, say $\widetilde{\mathbf{g}}_k(n)=[[\widetilde{\mathbf{g}}_{k,1}(n)]^T,\dots,[\widetilde{\mathbf{g}}_{k,s}(n)]^T,\dots,[\widetilde{\mathbf{g}}_{k,S_n}(n)]^T]^T$, where $\widetilde{\mathbf{g}}_{k,s}(n) \in \mathbb{R}^G$ is a $G_n$-entry vector representing the  spreading chips of $[\widehat{\mathbf{g}}^{\text{co}}_k(n)]_s$, given as 
\begin{equation}
\widetilde{\mathbf{g}}_{k,s}(n)=[\widehat{\mathbf{g}}^{\text{co}}_k(n)]_s\mathbf{C}_s(n), \quad \forall (s,  k).
\end{equation}
Note that for all $s\in \{1,2,\dots,S_n\}$, $\frac{1}{G_n} \sum_{\ell=1}^{G_n} [\mathbf{C}_s(n)]_\ell^2=1$ holds.

\subsection{Receiver Design}

The receiver design is illustrated in Fig. \ref{fig:SBtransceiver}(b) comprising three cascaded operations, i.e., spectrum de-spreading, signal de-normalization and zero padding.

\subsubsection{Spectrum De-spreading} 
The operation targets mining the desired gradient symbols hidden in the interference using the \textit{de-spreader} to be elaborated in the following.
Perfect synchronization between transmitters and receiver is assumed such that chip-level operations of spectrum de-spreading can be realized.
Considering round $n$, the received signals at the server are the superimposed waveform due to the simultaneous transmission of devices.
Let $\widetilde{\mathbf{y}}(n) \in \mathbb{R}^{S_n}$ denote the output vector of spectrum de-spreading. By introducing the truncated channel inversion in (\ref{Eq:TCI}), the $s$-th output element, say $[\widetilde{\mathbf{y}}(n)]_s$, is given as
\begin{equation} 
\begin{split}
     [\widetilde{\mathbf{y}}(n)]_s &= \frac{1}{G_n} \sum_{\ell=1}^{G_n} [\mathbf{C}_s(n)]_\ell \Re([\mathbf{y}_s(n)]_{\ell})\\
     &=\sum_{k\in \mathcal{K}(n)} \frac{\sqrt{P_0}}{G_n} \sum_{\ell=1}^{G_n}[\mathbf{C}_s(n)]^2_{\ell} [\widehat{\mathbf{g}}^{co}_k(n)]_{s}+[\widetilde{\mathbf{z}}(n)]_{s},\\
      &=\sqrt{P_0}\sum_{k\in \mathcal{K}(n)} [\widehat{\mathbf{g}}^{co}_k(n)]_{s}+[\widetilde{\mathbf{z}}(n)]_{s},
\end{split}
\end{equation}
where  $[\widetilde{\mathbf{z}}(n)]_s\sim \mathcal{N}(0,\frac{P_I}{G_n})$ is zero-mean Gaussian interference, whose power is inversely proportional to $G_n$ \cite{RN232}.

\subsubsection{Signal De-normalization}
This operation is performed to eliminate the impact of normalization and channel inversion to obtain the noisy averaged gradient symbols, denoted as $\widehat{\mathbf{y}}(n)\in\mathbb{R}^{S_n}$, given as
\begin{equation}
\begin{split}
       \widehat{\mathbf{y}}(n)
       &=\frac{V(n)}{\sqrt{P_0}|\mathcal{K}(n)|}\widetilde{\mathbf{y}}(n)+|\mathcal{K}(n)|M(n)\mathbf{1} \\
        &=\frac{1}{|\mathcal{K}(n)|}\sum_{k\in\mathcal{K}(n)}\mathbf{g}^{\text{co}}_k(n)+\frac{V(n)}{\sqrt{P_0}|\mathcal{K}(n)|}\widetilde{\mathbf{z}}(n).
\end{split}
\end{equation}

\subsubsection{Gradient Zero-padding}
To facilitate global-model updating, zero padding executes the inverse of pruning $\Psi^{-1}(\cdot):\mathbb{R}^{S_n} \rightarrow \mathbb{R}^D$ to restore the $D$-dimensional update by inserting zeros into the punctured dimensions. 
The zero-padded gradient, denoted as $\mathbf{y}^\prime(n)$, is represented as
\begin{equation}
\label{rece_signal}
    \begin{split}
       \mathbf{y}^\prime(n)=
       R\left(\frac{1}{|\mathcal{K}(n)|}\sum_{k\in\mathcal{K}(n)}\mathbf{g}_k(n)+\widehat{\mathbf{z}}(n)\right),
    \end{split}
\end{equation}
where  $\widehat{\mathbf{z}}(n)\in \mathbb{R}^D $ represents the interference vector distributed as $\mathcal{N}(0,\frac{V^2(n)P_I}{P_0|\mathcal{K}(n)|^2G_n}\mathbf{I}_D)$, and 
$R(\cdot):\mathbb{R}^D\times \Omega_n \rightarrow \mathbb{R}^D$ is the zero-padding operation;  Finally, devices update the global model using the gradient $\mathbf{y}^\prime(n)$ after its broadcasting from the server.


%

%

\section{Convergence Analysis of AirBreathing Federated Learning}
\label{Sec:ConvergenceAnalysis}
In this section, we analyze the convergence of AirBreathing FL. 
The results are useful for optimizing the spectrum breathing in the next section.

\subsection{Assumptions, Definitions, and Known Results}
For tractable analysis, some commonly used assumptions, definitions, and known results are provided below. First, we consider a strongly-convex loss function with bounded gradient estimates. These assumptions are commonly used in the literature (see, e.g., \cite{RN97,RN98,RN270}).

\begin{Assumption} \rm
\label{AS1}

The differentiable loss function $F(\cdot)$ is $c$-strongly convex, i.e., $\forall \mathbf{w}_1,\mathbf{w}_2 \in \mathbb{R}^D$,
    \begin{equation}
    \begin{split}
       F(\mathbf{w}_1) -F(\mathbf{w}_2) \geq \nabla F(\mathbf{w}_1)^T (\mathbf{w}_2-\mathbf{w}_1)+\frac{c}{2}\Vert \mathbf{w}_2-\mathbf{w}_1 \Vert^2.
          \end{split}
    \end{equation}
  
  \end{Assumption} 
  
  \begin{Assumption} \rm
  \label{AS2}
   Let $\mathbf{w}^*\in \mathbb{R}^D$ denote the optimality point of $F(\cdot)$.  For $\varepsilon>0$, there exists a success region indicating the convergence, defined as   $\mathcal{S} = \{\mathbf{w}|\Vert\mathbf{w}-\mathbf{w}^*\Vert^2\leq\varepsilon\}$.
  \end{Assumption}

\begin{Assumption}
\label{AS-gradient bound}
\rm
Local gradients $\mathbf{g}_k(n)$  are i.i.d. over devices $k\in \{1,2,\dots,K\}$ with unbiased estimate of the ground truth $\mathbf{g}(n)$ and bounded variance, i.e.,
        $\mathbb{E}[\mathbf{g}_k(n)] = \mathbf{g}(n),
        \mathbb{E}[\Vert \mathbf{g}_k(n)-\mathbf{g}(n)\Vert^2]\leq \sigma_g^2,
        \mathbb{E}[\Vert\mathbf{g}(n)\Vert^2]\leq \zeta^2,$
for all $(n,k)$, where $\sigma_g^2$ and $\zeta^2$ are constants.
\end{Assumption}

Next, we adopt the method of martingale-based convergence analysis in \cite{RN97}. To this end, a useful definition and some known results from \cite{RN97} are provided below.

\begin{Definition}[\cite{RN97}]\rm
\label{defsupermartingale}
A non-negative process $W_n(\mathbf{w}(n),\mathbf{w}(n-1),\dots,\mathbf{w}(0)): \mathbb{R}^{D\times (n+1)}\rightarrow\mathbb{R}$ is defined as a rate supermartingale with a scalar parameter $A$, called the horizon, if the following conditions hold.
\begin{enumerate}
    \item It must be a supermartingale \cite{fleming2011counting}, i.e., for any sequence $\mathbf{w}(n),\mathbf{w}(n-1),\dots,\mathbf{w}(0)$ and $\forall n \leq A$, 
    $\mathbb{E}[W_{n+1}(\mathbf{w}(n+1),\mathbf{w}(n),\dots,\mathbf{w}(0))] \leq W_n(\mathbf{w}(n),\dots,\mathbf{w}(0))$ holds.
   

    \item For all rounds $N\leq A$ and  for any sequence $\mathbf{w}(n),\mathbf{w}(n-1),\dots,\mathbf{w}(0)$, if the algorithm has not converged into the success region by $N$ (i.e., $\mathbf{w}(n)\notin \mathcal{S}, \forall n\leq N$),
 $W_N(\mathbf{w}(N),\mathbf{w}(N-1),\dots,\mathbf{w}(0))\geq N$ holds.
\end{enumerate}
\end{Definition}

\begin{Lemma} [\cite{RN97}, Lemma 1] \rm
\label{W_n}
Consider an FL system updating as in (\ref{gsgd}) with a learning rate $\eta<2c\varepsilon \mathbb{G}^2$.  If the algorithm has not converged by round $n$, the process defined as
 \begin{equation}
\begin{split}
& W_n({\mathbf{w}}(n), \ldots, {\mathbf{w}}(0))\\
\triangleq &  \frac{\varepsilon}{2 \eta  c \varepsilon-\eta^{2}\mathbb{G}^2} \log \left(\frac{e\left\Vert\mathbf{w}(n)-{\mathbf{w}}^{*}\right\Vert^{2}}{\varepsilon}\right)+n, 
\end{split}
\label{W_t expression}
\end{equation}
 is a rate supermartingale with horizon $A=\infty$, where $\mathbb{G}^2 \geq \zeta^2+\sigma_g^2$ is the upper bound of the squared norm of aggregated gradients.
 Under Assumptions \ref{AS1}-\ref{AS2}, $W_n({\mathbf{w}}(n), \ldots, {\mathbf{w}}(0))$ is also $H$-Lipschitz smooth in the first coordinate, with $H=2\sqrt{\varepsilon}(2\eta c \varepsilon-\eta^2 \mathbb{G}^2)^{-1}$. In other words, for any $n\geq 1,\mathbf{w}_1,\mathbf{w}_2\in \mathbb{R}^{D}$ and any sequence $\mathbf{w}(n-1), \ldots, \mathbf{w}(0)$, it satisfies
\begin{equation}
\begin{split}
\label{H-Lip}
    \Vert W_n(\mathbf{w}_1,\mathbf{w}(n-1), \ldots, \mathbf{w}(0))-  W_n(\mathbf{w}_2,\mathbf{w}(n-1), \\ \ldots, \mathbf{w}(0))\Vert  \leq  H \Vert \mathbf{w}_1-\mathbf{w}_2\Vert.
    \end{split}
\end{equation}

\end{Lemma}

Intuitively, the rate supermaringale  represents the level of satisfaction for model weights $\mathbf{w}(n),\mathbf{w}(n-1),\dots,\mathbf{w}(0)$ over $n+1$ rounds.
For Definition \ref{defsupermartingale}, some intuition into the preceding assumptions are as follows.
First,  condition 1)  reflects the fact that  obtained model weights are more satisfactory as they approach the optimality point. 
Second, as specified in  condition 2), the satisfaction is reduced if the algorithm is executed for many rounds without convergence.
FL updating as in (\ref{gsgd}) is considered as the vanilla SGD satisfying the properties of rate supermartingale.
It is a commonly used analytical method in the convergence analysis of SGD \cite{RN97,RN98,RN270}.

\subsection{Convergence Analysis}
Based on the preceding assumptions,  we further develop the mentioned rate-supermartingale approach to study the convergence of AirBreathing FL.
The new approach is able to account for channel fading, and the system operations such as AirComp and spread spectrum.
Specifically, several useful intermediate results are obtained as shown in the following lemmas.

We first upper bound the gap between the vanilla SGD and AirBreathing FL using the results on the AirComp error (see Lemmas \ref{MSEG}  and \ref{MNEbound}).
Next, based on Lemma \ref{MNEbound}, a supermartingale for AirBreathing FL is constructed in Lemma \ref{U_n}.
Furthermore, the upper bound of the convergence rate is derived using the theory of martingale as shown in Lemma \ref{The_1}.

\begin{Lemma}\rm
\label{MSEG}

The AirComp error defined in (\ref{MSE_metric}) for round $n$, can be expressed as the sum of the gradient-pruning error and interference-induced error:
\begin{equation}
\label{MSE_R}
\begin{split}
    \textup{MSE}(n)=\underbrace{ \left(1-\gamma_n\right)  \mathbb{E}[\alpha^2(n)] }_{\textup{gradient-pruning error}} + \underbrace{ \frac{\gamma_n D  P_I}{ G_n P_0}\mathbb{E}\left[\frac{V^2(n)}{|\mathcal{K}(n)|^2}\right]}_{\textup{interference-induced error}},
\end{split}
\end{equation}
where  $\gamma_n=\frac{S_n}{D}$ represents the pruning ratio in round $n$, and $\alpha^2(n)$ is defined as
\begin{equation}
\label{MSE(n)}
    \begin{split}
         \alpha^2(n)=\left\Vert\frac{1}{|\mathcal{K}(n)|}\sum_{k\in\mathcal{K}(n)}\mathbf{g}_k(n)\right\Vert^2.
    \end{split}
\end{equation}

\noindent  Proof.  \textup{See Appendix \ref{Proof of MSEG}.}

\end{Lemma}

\begin{Lemma}\rm
\label{MNEbound}

Considering round $n$, the gap between vanilla SGD and AirBreathing FL is defined as the expected difference between the update of vanilla SGD, namely $\frac{1}{K}\sum_{k=1}^K\mathbf{g}_{k}(n)$, and spectrum breathing, namely $\mathbf{y}^\prime(n)$. The gap can be bounded as
\begin{equation}
   \mathbb{E}\left[\left\Vert \frac{1}{K}\sum_{k=1}^K\mathbf{g}_{k}(n)- \mathbf{y}^\prime(n) \right\Vert\right]\leq u(n),
\end{equation}
where $u(n)$ is defined as
\begin{equation}
\label{u(n)}
    u(n)= \frac{2-\xi_a}{K\xi_a}\sigma^2_g + \sqrt{\textup{MSE}(n)},
\end{equation}
 where $\xi_a$ and $\sigma_g^2$ is the activation probability in (\ref{active-prob}) and gradient variance in Assumption \ref{AS-gradient bound}, respectively.
The first and second terms of the bound result from 1) the fading channel and gradient randomness  and 2) the AirComp error in (\ref{MSE_R}), respectively.

\noindent  Proof. \textup{See Appendix \ref{proof of MNEbound}. }
\end{Lemma}

The process defined in Lamma \ref{MNEbound}, $\{u(n)\}$, serves as indicators of performance loss caused by the air interface and is hence termed \emph{propagation-loss process}.
To this end, we use the result in Lamma \ref{MNEbound} to define a new stochastic process pertaining to spectrum breathing and  show it to be a supermartingale. The details are as follows.

\begin{Lemma}\rm
\label{U_n}

Define a stochastic process, $\{U_n\}$, as 
\begin{equation}
    U_n(\mathbf{w}(n),\dots,\mathbf{w}(0))\triangleq W_n(\mathbf{w}(n),\dots,\mathbf{w}(0))-\eta H\sum_{i=0}^{n-1}u(i),
\end{equation}
for $\forall i \leq n$ and $\mathbf{w}(i)\notin \mathcal{S}$, $\{U_n\}$ is a supermartingale process.

\noindent Proof. \textup{See Appendix \ref{Proof of U_n}.}
\end{Lemma}

Note that, $U_n$ has a negative term, which is a function of the propagation-loss process, removes from the model under training
the effect of the air interface. Thereby, the result in Lemma \ref{U_n} facilitates the use of supermartingale theory to quantify the convergence probability of AirBreathing FL as shown below.



\begin{Lemma}    \rm
\label{The_1}
Consider $N$ rounds and AirBreathing FL for minimizing the loss function $F(\mathbf{w})$. 
If the learning rate satisfies
 \begin{equation}
 \label{lr}
     \eta<\frac{2\sqrt{\varepsilon}\left(c \sqrt{\varepsilon} N- \sum_{n=0}^{N-1} u(n)\right)}{N \mathbb{G}^{2}},
 \end{equation}
 the event of failing to converge to the success region, denoted as $F_N$, has a probability bounded as
 \begin{equation}
 \label{PrFT}
     \begin{split}
         \textup{Pr}\{F_N\}\leq
         \frac{\varepsilon \log\left(e\Vert \mathbf{w}(0)-\mathbf{w}^*\Vert^2 \varepsilon^{-1}\right)}{(2\eta c \varepsilon-\eta^2\mathbb{G}^2)N  -2\eta \sqrt{\varepsilon}\sum_{n=0}^{N-1}u(n)}.
     \end{split}
 \end{equation}
 where $\mathbb{G}^2$ is the upper bound of aggregated gradient defined in Lemma \ref{W_n}.

\noindent  Proof: \textup{See Appendix \ref{proof of The_1}}. 
\end{Lemma}



\begin{Definition}[Breathing Depth] \rm
In the considered scenario of constrained bandwidth-and-latency, it is necessary to fix the product of processing gain and pruning ratio: $G_n\gamma_n=1$. Under this constraint, the tradeoff between spread spectrum and gradient pruning can be regulated by the processing gain $G_n=\frac{1}{\gamma_n}$. To be more instructive, it is renamed the \emph{Breathing Depth}, that is the most important control variable of spectrum breathing.
    
\end{Definition}

Substituting $G_n=\frac{1}{\gamma_n}$ into the result in Lemma \ref{The_1} yields the following main result.

\begin{Theorem}\rm
\label{Cor_1}

Consider AirBreathing FL with breathing depths $\{G_n\}$ and $N$ rounds. If the learning rate satisfies (\ref{lr}), 
 the probability of failing to converge to the success region is bounded as
 \begin{equation}
 \label{PrFT_Cor_1}
     \begin{split}
         \textup{Pr}\{F_N\}\leq 
              \frac{\varepsilon \log\left(e\Vert \mathbf{w}(0)-\mathbf{w}^*\Vert^2 \varepsilon^{-1}\right)}{\left(2 c \varepsilon-\eta\mathbb{G}^2-2\sqrt{\frac{(2-\xi_a)\varepsilon}{M\xi_a}}\sigma_g \right)\eta N  -2\eta \sqrt{\varepsilon} \beta_\Sigma}.
     \end{split}
 \end{equation}
 Here $\beta_{\Sigma}=\sum_{n= 0} ^{N-1}\sqrt{\beta_n(G_n)}$ is a sum of error terms where each term  $\beta_n(G_n)$ represents the air-interface error in round $n$, given as
 \begin{equation}
 \label{air-interface error}
   \beta_n(G_n) = \underbrace{ \left(1-\frac{1}{G_n}\right)  \mathbb{E}[\alpha^2(n)] }_{\textup{gradient-pruning error}} + \underbrace{ \frac{ D  P_I}{ G^2_n P_0}\mathbb{E}\left[\frac{V^2(n)}{|\mathcal{K}(n)|^2}\right]}_{\textup{interference-induced error}}.
 \end{equation}

\end{Theorem}

Consider the air-interface error term in (\ref{air-interface error}). One can see that increasing the breathing depth, $G_n$, corresponds to decreasing the pruning ratio and thus causes the pruning error to grow. On the other hand, increasing $G_n$ enhances the process gain and thereby reduces interference perturbation (and  its corresponding error term). The above tradeoff suggests the need of optimizing $\{G_n\}$, which is the topic of the next section.

Comparing Theorem 1 to the convergence analysis in related works \cite{RN133,RN270}, our results have two main differences:
First,  our results reflect the effect of spectrum breathing depth, $G_n$, in (\ref{air-interface error}), which does not exist in prior work.
When $G_n=1$ (no breathing, pruning), (\ref{air-interface error}) reduces to the mean squared norm of the introduced noise in \cite[(18)]{RN133}.
Second, the effect of fading channels on convergence is characterized in (\ref{air-interface error}) by the term  $\frac{1}{|\mathcal{K}(n)|^2}$,  while the $|\mathcal{K}(n)|$ in \cite{RN270} is assumed to be constant.


%

\section{Optimization of Spectrum Breathing} \label{Sec:optmization}

In this section, the results from the preceding convergence analysis are applied to the optimization of the breathing depth of AirBreathing FL. To enhance convergence, Theorem \ref{Cor_1} imposes the need of minimizing (\ref{air-interface error}).
 Before that, the assumptions on known and unknown parameters are specified as follows. 
The predefined parameters, i.e., model size $D$ and receive SIR $\frac{P_0}{P_I}$, are assumed to be known.
Let \emph{Gradient State Information} (GSI) refer to the statistical parameters of the stochastic gradient in the current round, namely $\alpha(n)$ and $V(n)$ in (\ref{air-interface error}), which are not accessible but can be estimated at each round using local gradients.
Moreover, let CSI refer to the channel-dependent number of active devices in the current round, namely $|\mathcal{K}(n)|$.
It is known to the server by assuming perfect channel estimation over rounds. 
Then we consider the optimization in two cases: (1) without GSI and CSI, and (2) with GSI and CSI at the server.

\subsection{Breathing Depth Optimization without GSI and CSI Feedback}
\label{Fixed PG}

Without GSI and CSI  feedback, we deploy fixed breathing depth  for all rounds, i.e., $G_n=G,\forall n$.
Consider the term $\beta_\Sigma$ in Theorem \ref{Cor_1}, which is the only term related to the breathing depth, $G$. Then $G$ is optimized to minimize $\beta_\Sigma$, thereby accelerating convergence. The difficulty of such optimization lies in the lack of the required GSI and CSI. To address the issue, we resort to minimizing an upper bound on $\beta_\Sigma$ that requires no such information.

\begin{Lemma}\rm
    \label{alpha+V}
    
    Consider gradient $\mathbf{g}_k(n), k \in \mathcal{K}(n)$. There exists a positive constant $\Gamma(n)$, satisfying $\Gamma(n)\geq  \frac{1}{D}(\Vert\mathbf{g}(n)\Vert^2+\sigma^2_g)$, such that $\beta_\Sigma$ is upper bounded as
 \begin{equation}
     \beta_\Sigma\leq \beta_F(G)  \sum_{n=0} ^{N-1}\sqrt{D\Gamma(n)},
 \end{equation}
 where $\beta_F(G)$ is a function of $G$, given as
\begin{equation}
    \beta_F(G)=\sqrt{ 1-\frac{1}{G}+\frac{6P_I}{G^2K^2\xi_a^2P_0}}.
\end{equation}

 \noindent Proof. \textup{See Appendix \ref{Proof of alpha+V}}.
\end{Lemma}
We next formulate the optimization problem
\begin{equation}
\label{fixed PG optim}
    \begin{split}
        \min_G &\quad  \beta_F(G)  \\
         \text{s.t.}& \quad G \in \{1,2,\dots,D\}.
    \end{split}
\end{equation}

Problem (\ref{fixed PG optim}) can be solved by integer relaxation as follows. Given a continuous variable $x>0$, setting $\nabla_x\beta_F(x)=0 $  yields the optimal solution $x^*=\frac{12 P_I}{P_0 K^2  \xi^2_a}$. Thus, for the discrete function $\beta_F(G),\forall G\in \{1,2,\dots, D\}$, the fixed breathing depth, denoted $G^*$, can be obtained approximately by
\begin{equation}
\label{optimal_G_randk}
    G^*=
    \begin{cases}
    1,&x^*<1,\\
\left\lfloor x^*\right\rfloor_{\beta_F(G)}, &  1\leq x^*\leq D,\\
    D,&x^*> D,
    \end{cases}
\end{equation}
where  $\lfloor \widehat{x} \rfloor_{\beta_F(x)}$ is equal to $\lfloor \widehat{x} \rfloor$ if $ \beta_F(\lfloor \widehat{x} \rfloor)\leq \beta_F(\lceil \widehat{x} \rceil) $, and is otherwise equal to $\lceil \widehat{x} \rceil$.

In the above results, $G^*$ is a monotonous decreasing function of  the receive SIR $\frac{P_0}{P_I}$. It implies that for a low receive SIR, we allocate more bandwidth resources for interference suppression to guarantee convergence at cost of more aggressive gradient compression.  On the other hand, for a high receive SIR, the spectrum-breathing control favours uploading as many gradient dimensions as possible to attain faster convergence. 
Increasing the expected number of active devices $K\xi_a$ directly enhances the received signal power, which suppresses the interference by aggregation gain and hence reduces the need of interference suppression via spectrum spreading.

\subsection{Breathing Depth Optimization with GSI and CSI Feedback}
\label{Adap PG}

Given GSI and CSI feedback, the breathing depth can be adapted over rounds and hence re-denoted as $G_n$ for round $n$. The optimization criteria is to minimize the estimate of the relevant term, $\beta_n(G_n)$, of the successful convergence probability in the Theorem \ref{Cor_1}. Let the estimate be denoted as $\widehat{\beta}_n(G_n)$:
\begin{equation}
\label{widehat{beta}_n}
   \widehat{\beta}_n(G_n)=\left( 1-\frac{1}{G_n}\right)\widehat{\alpha}^2(n)+\frac{DP_I \widehat{V}^2(n)}{G_n^2 P_0|\mathcal{K}(n)|^2 },
\end{equation}
where $\widehat{\alpha}(n)$ and $\widehat{V}(n)$ are estimates of  $\alpha(n)$ and $V(n)$, respectively, while the number of active devices, $|\mathcal{K}(n)|$, is perfectly known at the server from CSI. Based on feedback statistics of local gradients, the estimation is similar to that in \cite{RN10} given as
\begin{equation}
\label{estimate_alphat}
\begin{split}
     \widehat{\alpha}^2(n) &= \frac{1}{|\mathcal{K}(n)|}\sum_{k \in \mathcal{K}(n)}\Vert\mathbf{g}_k(n) \Vert^2, \\
     \widehat{V}^2(n) & =\frac{1}{|\mathcal{K}(n)|}\sum_{k \in \mathcal{K}(n)}\widehat{V}_k^2(n),
\end{split}
\end{equation}
where $\widehat{V}_k^2(n)$ is the local gradient variance:
\begin{equation}
\label{local_variance}
\begin{split}  
   \widehat{V}_k^2(n) =& \frac{1}{D}\sum_{d=1}^D \left( [\mathbf{g}_k(n)]_d-\frac{1}{D} \sum_{d=1}^D [\mathbf{g}_k(n)]_d \right)^2.
\end{split}
\end{equation}
Consider an arbitrary round $n$ of AirBreathing FL. Based on (\ref{widehat{beta}_n}) and (\ref{estimate_alphat}), the problem of optimizing the breathing depth can be formulated as 
\begin{equation}
\label{P_Gt_each}
    \begin{split}
        \min_{G_n}  & \quad\widehat{\beta}_n(G_n)\\
        \text{s.t.}& \quad G_n \in \{1,2,\dots,D\}, \\
         &\quad \forall n\in\{0,1,\dots,N-1\}.
    \end{split}
\end{equation}
Again, by integer relaxation of $G_n$ into $x>0$, $\nabla_x \widehat{\beta}_n(x)=0$  yields the optimal solution $x_n^*=\frac{2 P_I D \widehat{V}^2(n)}{P_0|\mathcal{K}(n)|^2\widehat{\alpha}^2(n)}$ such that  $\widehat{\beta}_n(x_n^*)$ is the minimum.
Then an approximate of the adaptive breathing depth is given as 
\begin{equation}
\label{optimal_G}
    G_n^*=
    \begin{cases}
    1,& x_n^* <1,\\
    \left\lfloor x_n^*\right\rfloor_{\widehat{\beta}_n(G_n)}, &  1\leq x_n^*\leq D,\\
    D, & x_n^* \geq D.
    \end{cases}
\end{equation}

\begin{algorithm}[t]
 \label{Alg:Gt}
 \caption{Adaptive Breathing Depth Protocol}
 \begin{algorithmic}[1]
 \renewcommand{\algorithmicrequire}{\textbf{Input:}}
  \REQUIRE  Receive SIR $P_0/P_I$, Model size $D$;
\STATE{Initialisation} : $\mathbf{w}(0)$ in all devices;
  \FOR  {Round: $n = 0$ to $N$}
  \FOR{each device $k\in \mathcal{K}(n)$ in parallel}
  \STATE Computes $\mathbf{g}_k(n)$ via (\ref{grad});
  \STATE Computes $\Vert \mathbf{g}_k(n) \Vert^2$ ;
  \STATE Computes $\widehat{V}^2_k(n)$ via (\ref{local_variance});
  \STATE Uploads $\Vert \mathbf{g}_k(n) \Vert^2$ and $\widehat{V}^2_k(n)$ to server;
    \ENDFOR
    \STATE Server estimates $\widehat{\alpha}^2(n)$ and $\widehat{V}^2(n)$ via (\ref{estimate_alphat});
    \STATE Server computes adaptive breathing depth $G_n^*$ via (\ref{optimal_G});
    \STATE Server generates selected index set $\psi_n$ and set of PN sequence $\mathcal{C}(n)$ w.r.t. $G_n^*$;
    \STATE Server broadcasts $\psi_n$ and $\mathcal{C}(n)$ to all devices;
    \STATE \textbf{Spectrum Breathing Process} returns $\mathbf{w}(n+1)$
  \ENDFOR
  \RETURN $\mathbf{w}(n+1)$
 \end{algorithmic} 
 \end{algorithm}

 Note that the adaptive breathing depth is a clipping function of $x_n^*$, truncated by the smallest and largest achievable value.
For the general case, $1\leq x_n^*\leq D$,  $G_n^*$ is found to be 
inversely proportional to the receive SIR $\frac{P_0}{P_I}$, number of active devices $|\mathcal{K}(n)|^2$ and estimate of gradient squared norm $\widehat{\alpha}^2(n)$. 
Enlarging these parameters reduces the impact of interference on convergence so that breathing depth decreases accordingly. On the other hand, the breathing depth increases with the rise of  gradient-variance estimate due to the scaling of interference in de-normalization.
The resultant protocol for adaptive spectrum breathing is summarized in Algorithm \ref{Alg:Gt}.
The spectrum-breathing transceiver has a linear $\mathcal{O}(D)$ complexity per round. This is notably more efficient than the $\mathcal{O}(D^2)$ complexity of importance-aware pruning.

\section{Experimental Results}\label{Sec:experiments}

In this section, the preceding fixed and adaptive breathing depth protocols are simulated. Based on this, we evaluate the performance of AirBreathing FL by comparing it with six benchmarks to be specified below.

\subsection{Experimental Settings}
The default experimental settings are as follows unless specified otherwise. 
\begin{itemize}
    \item \textbf{Communication Settings:} 
    We consider an AirBreathing FL system comprising one server and 10 devices.
    In each round,  the PN sequence shared by devices is generated by having i.i.d. chips following the unbiased Bernoulli distribution; 
    the sequence is varied over rounds.
    Each chip  spans unit time and a transmitted gradient coefficient occupies $G_n$ chips with $G_n$ being the breathing depth.
    The interference at the server's receiver is modelled as a sequence of i.i.d Gaussian symbols.
    Assuming Rayleigh fading, all channel coefficients are modelled as $\mathcal{CN}(0,1)$ random variables.
    Consider the scenario of strong interference. The devices' fixed transmission power and the interference power are set such the expected receive SIR is $-23$dB, which can be enhanced by aggregation and spectrum de-spreading in AirBreathing.
    Finally, the threshold of truncated channel inversion is set as $G_{th}=0.2$ and the resultant activation probability of each device is $\xi_a=0.82$.

    \item \textbf{Learning Settings:} \
    We consider the learning task of   handwritten digit classification using the popular MNIST dataset. MNIST is widely considered for evaluating gradient-pruning schemes \cite{RN193,RN270} and de-noising techniques  \cite{RN34,RN106}.
   MNIST comprises two separate data subsets for training (60000 samples)
and validation (10000 samples), respectively. To model non-i.i.d data at devices, each of which comprises 3000 randomly drawn samples of one class from the training dataset.
Two randomly chosen shards with different labels are assigned to each device.
The task is to train a CNN model having 21,840 parameters. The model consists of two $5 \times 5$ convolutional layers with ReLU activation (with 10, 20 channels, respectively), and the ensuing $2\times2$ max pooling, a fully connected layer with 50 units and ReLU activation, and a final softmax output layer.
During the training process, the learning performance is evaluated by the \emph{validation accuracy}, which is defined as the classification accuracy on the validation dataset comprising 10000 samples.
    Furthermore, the pruning for spectrum contraction is executed on model weights (99.8\% of all parameters) but not bias to avoid divergence. 
\end{itemize}

Six benchmarking schemes with their legends in brackets are described below.
 \begin{itemize}
    \item \textbf{Ideal Case}: The ideal FL system without pruning and channel distortion.
     \item \textbf{No Spectrum Breathing (No SB)}: This is equivalent to AirBreathing FL with $G_n=1, \forall n$. The resultant system is exposed to strong interference.
     \item \textbf{Pruning without Spectrum Spreading}:
     Gradients are pruned randomly with a fixed ratio, $\gamma$, and uploaded without spread spectrum. This results in the exposure of pruned gradients to strong interference.
     \item \textbf{Convergent OTA FL (COTAF) \cite{RN133}}: 
     Given the same power constraint and interference, time-varying scalar precoding (equivalent to power control) with full-dimensional gradient uploading is simulated by accounting for the gradually decreasing squared norm of gradient over rounds. Note that COTAF requires offline simulation to estimate the scalar precoder, while AirBreathing FL does not need this as a result of online estimation from GSI feedback with negligible communication overhead.
     \item \textbf{Optimal Fixed Breathing Depth (Optimal fixed BD)}: The optimal fixed breathing depth is obtained using an exhaustive search, as opposed to using the closed-form result in (\ref{optimal_G_randk}).
     \item \textbf{AirBreathing with Importance-aware Pruning (AirBreathing with IP)}:
     Random pruning is replaced with importance-aware pruning, namely pruning gradient coefficients with the smallest magnitudes.
     The difficulty in its implementation is overcome by alternating rounds of 1) full-gradient uploading to allow the devices to select from aggregated gradient coefficients an index subset to prune in the next round and 2) using the subset to perform importance-aware pruning at devices assuming temporal correlation in gradients \cite{RN192}.

 \end{itemize}

\subsection{Performance of AirBreathing FL}


\begin{figure}[t!]
    \centering
    \subfigure[Performance of AirBreathing FL]{    \includegraphics[width=0.8\columnwidth]{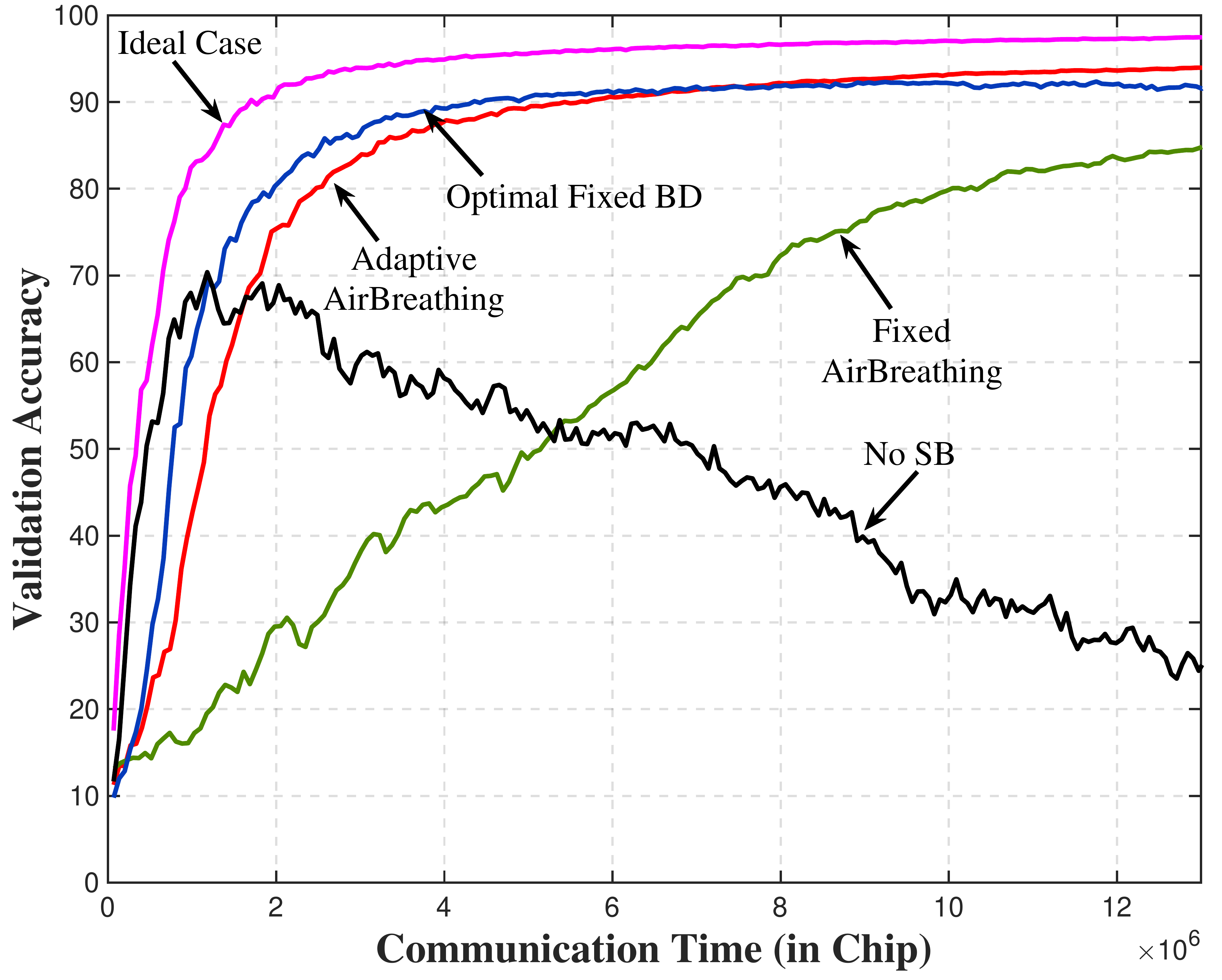}}
    \subfigure[Performance of pruning without spreading]{
    \includegraphics[width=0.8\columnwidth]{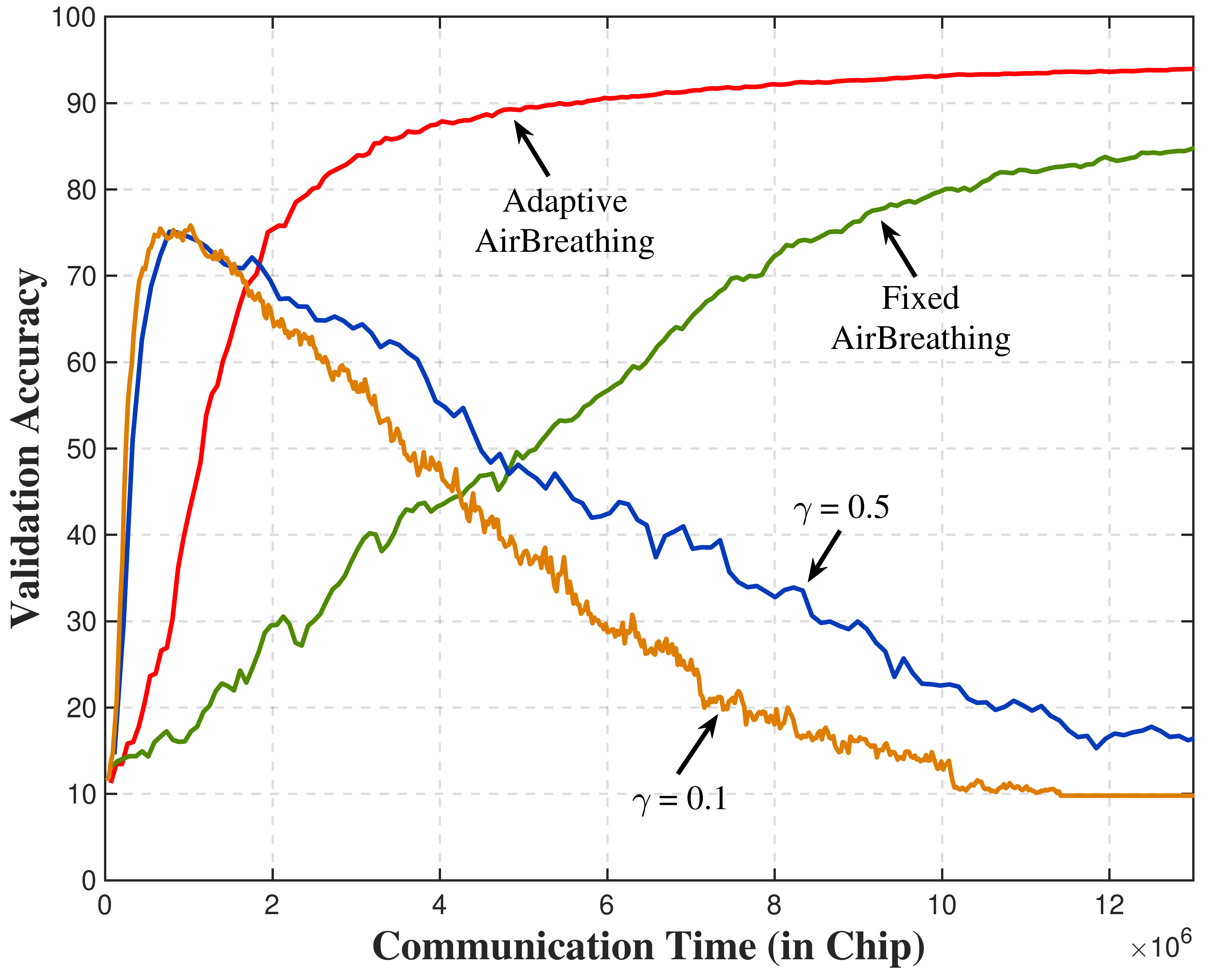}}
    \caption{(a) Performance comparison between AirBreathing FL (in both the cases of fixed and adaptive breathing depth) and benchmarking schemes; (b) Comparison between pruning without spreading ($\gamma$ = {0.5, 0.1}) and AirBreathing FL.}
    \label{fig:fixed_adap-only_pruning}\vspace{-3mm}
\end{figure}

The learning performance of AirBreathing FL is compared with the benchmarking schemes. For spectrum breathing, both the schemes of fixed and adaptive breathing depth are considered.
The curves of validation accuracy versus communication time are plotted in Fig. \ref{fig:fixed_adap-only_pruning} (a).
Several observations can be made.
First, AirBreathing in both the cases of fixed and adaptive breathing depth achieves convergence.
This demonstrates its effectiveness in coping with strong interference.
On the contrary, FL without spectrum breathing, which suffers from strong interference, fails to converge.
Second, there exists a substantial performance gap between the scheme of fixed breathing depth from the ideal case due to approximation in the former design to obtain the closed-form result.
Thus the gap is largely removed using the proposed scheme of adaptive breathing depth.
Finally, AirBreathing with adaptive depth is observed to approach the ideal case within a reasonable performance gap. In particular, the converged accuracy for the former is 96.2\% and 94.6\% for the latter.

\begin{figure}[t!]
    \centering
    \subfigure[Performance of COTAF]{
    \includegraphics[width=0.8\columnwidth]{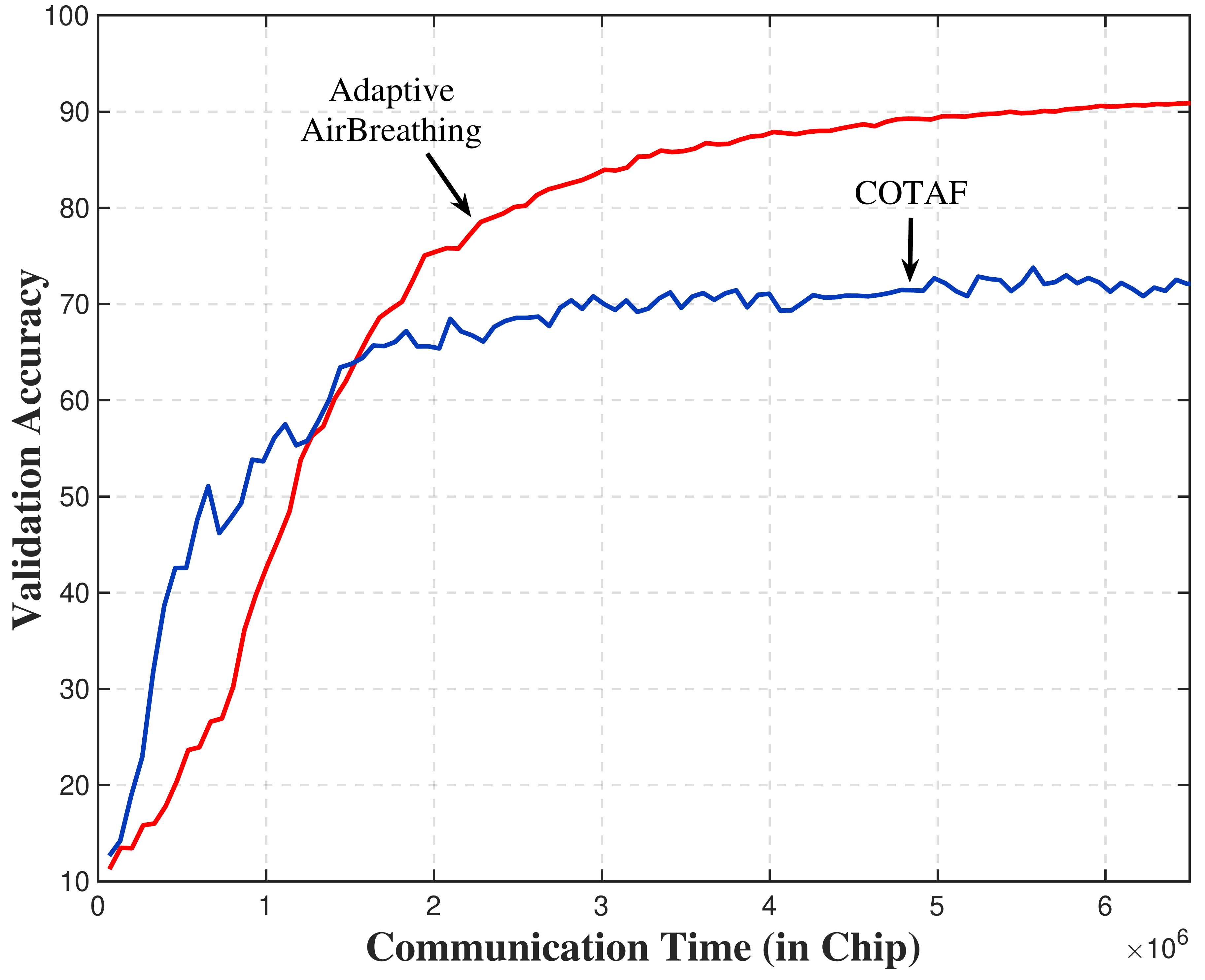}}
    \subfigure[Performance of two pruning schemes]{
    \includegraphics[width=0.8\columnwidth]{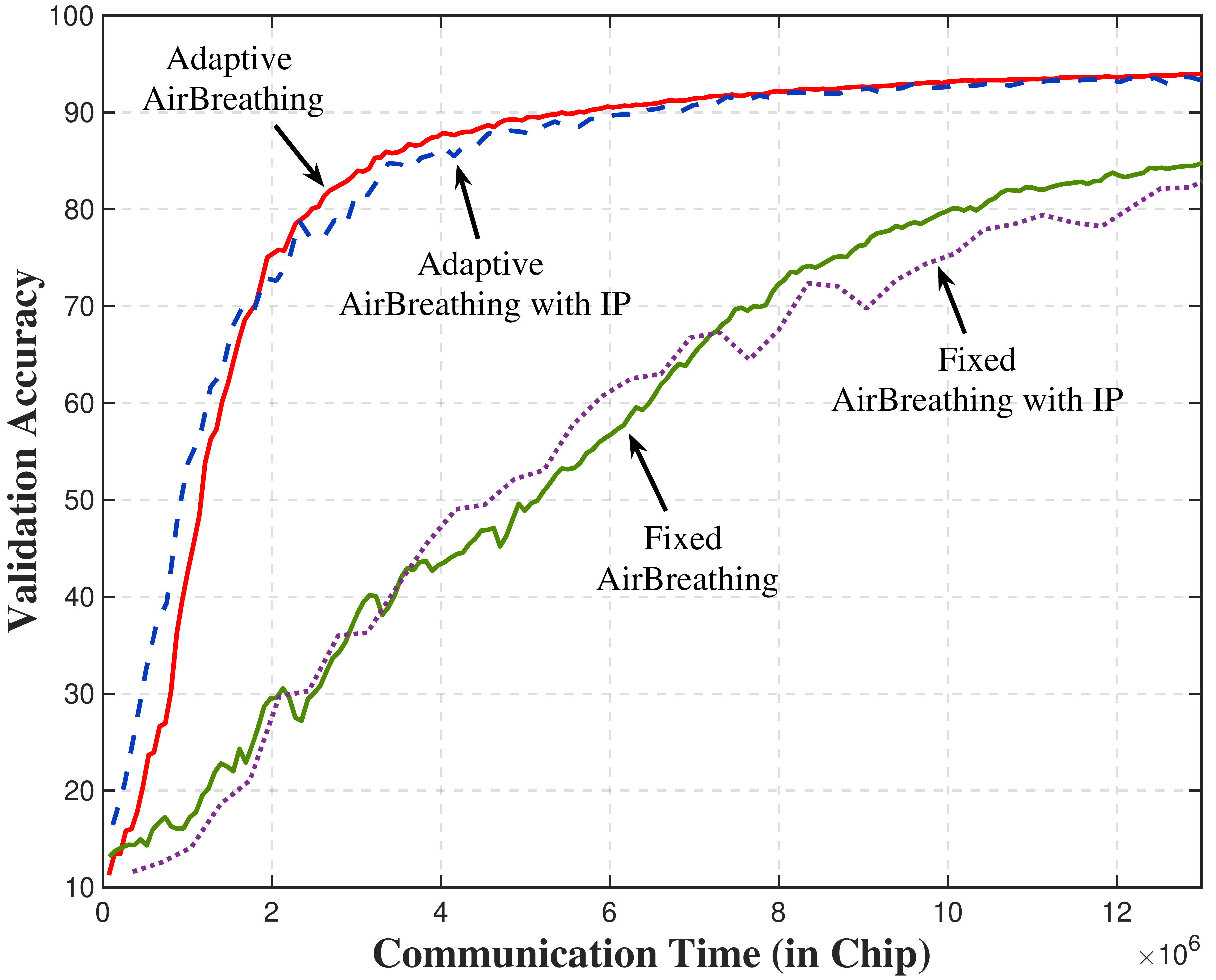}}
    \caption{(a) Performance comparison between COTAF and AirBreathing FL; (b) Comparison between random pruning and importance-aware pruning for AirBreathing FL.}
    \label{COTAF-importance}
    \vspace{-3mm}
\end{figure}

Fig. \ref{fig:fixed_adap-only_pruning} (b) compares  the performance of pruning without spectrum spreading and AirBreathing FL.
One can see the former  has a rapid increase in accuracy at the beginning, as a result from a higher communication rate in the absence of spectrum spreading.
However, the corruption of gradients by interference eventually takes its toll and leads to unsuccessful learning. In contrast, despite a slower learning speed initially, AirBreathing FL ensures steady increase in accuracy to achieve convergence.

Fig. \ref{COTAF-importance} (a) compares the performance between COTAF and adaptive AirBreathing FL in coping with strong interference. Several observations can be made.
When subjected to the same power constraints, COTAF avoids divergence by gradually increasing transmit power. Despite this, COTAF still struggles to achieve a high accuracy due to the discussed limitation of power control in suppressing interference.
In contrast, adaptive AirBreathing FL achieves a significantly higher accuracy after convergence, albeit with a slower learning speed at the beginning, thanks to its interference-suppression capability.

\subsection{Random Pruning versus Importance-aware Pruning}

In Fig. \ref{COTAF-importance} (b), we compare the learning performance of AirBreathing FL using the proposed random pruning with that of the benchmarking scheme using importance-aware pruning.
The main observation is that the latter does not yield any performance gain over the proposed scheme.
The reasons are two drawbacks of the importance-aware scheme.
First, the full-gradient uploading in every other round in the benchmarking scheme increases communication overhead.
Second, the pruned gradient coefficients in a round are selected based on those in the preceding round, resulting in inaccurate choices as gradients vary over rounds.


\subsection{Effects of Network Parameters}

\begin{figure}[t!]
    \centering
    \subfigure[Effects of number of devices]{
    \includegraphics[width=0.8\columnwidth]{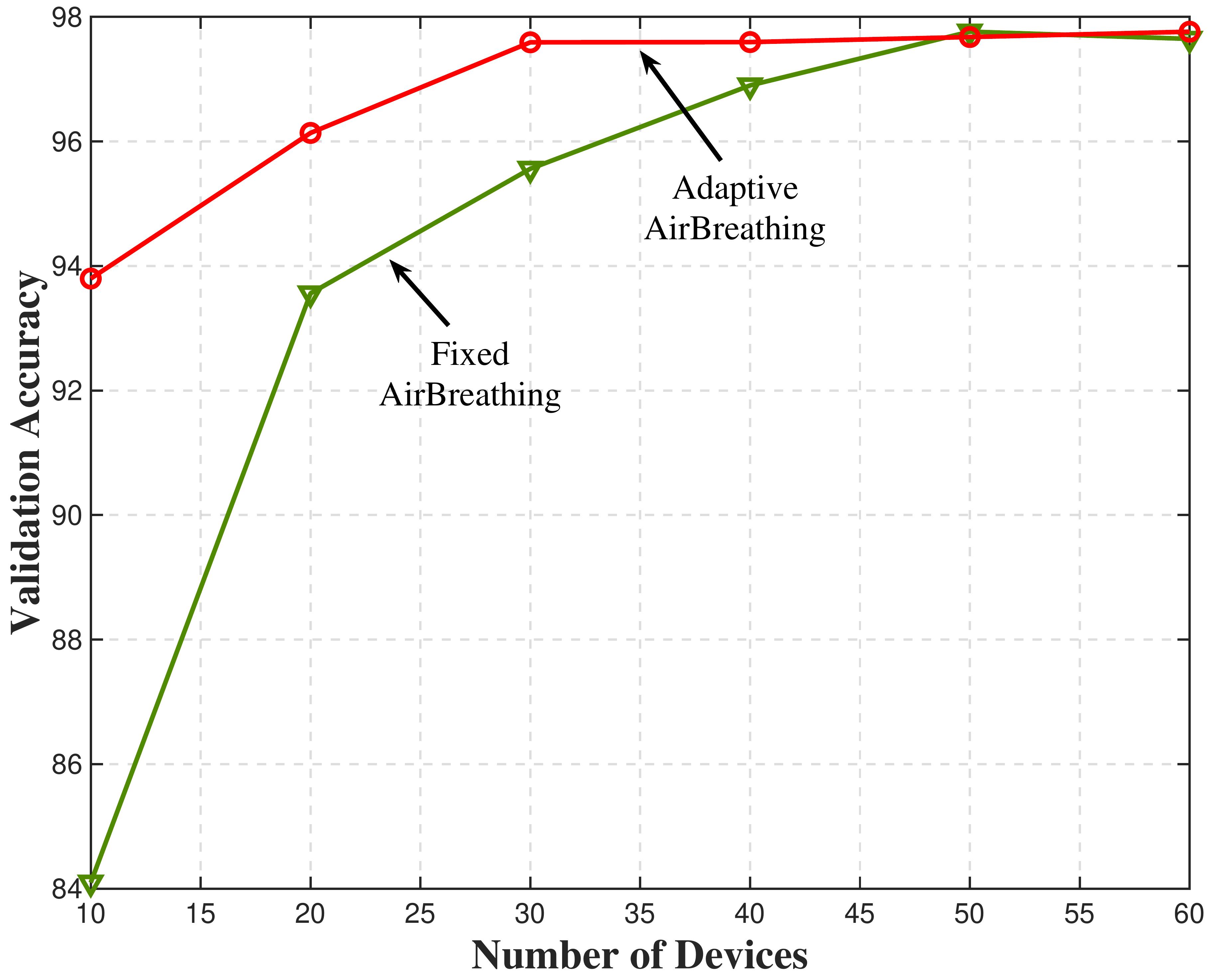}}
    \subfigure[Effects of receive SIR]{
    \includegraphics[width=0.8\columnwidth]{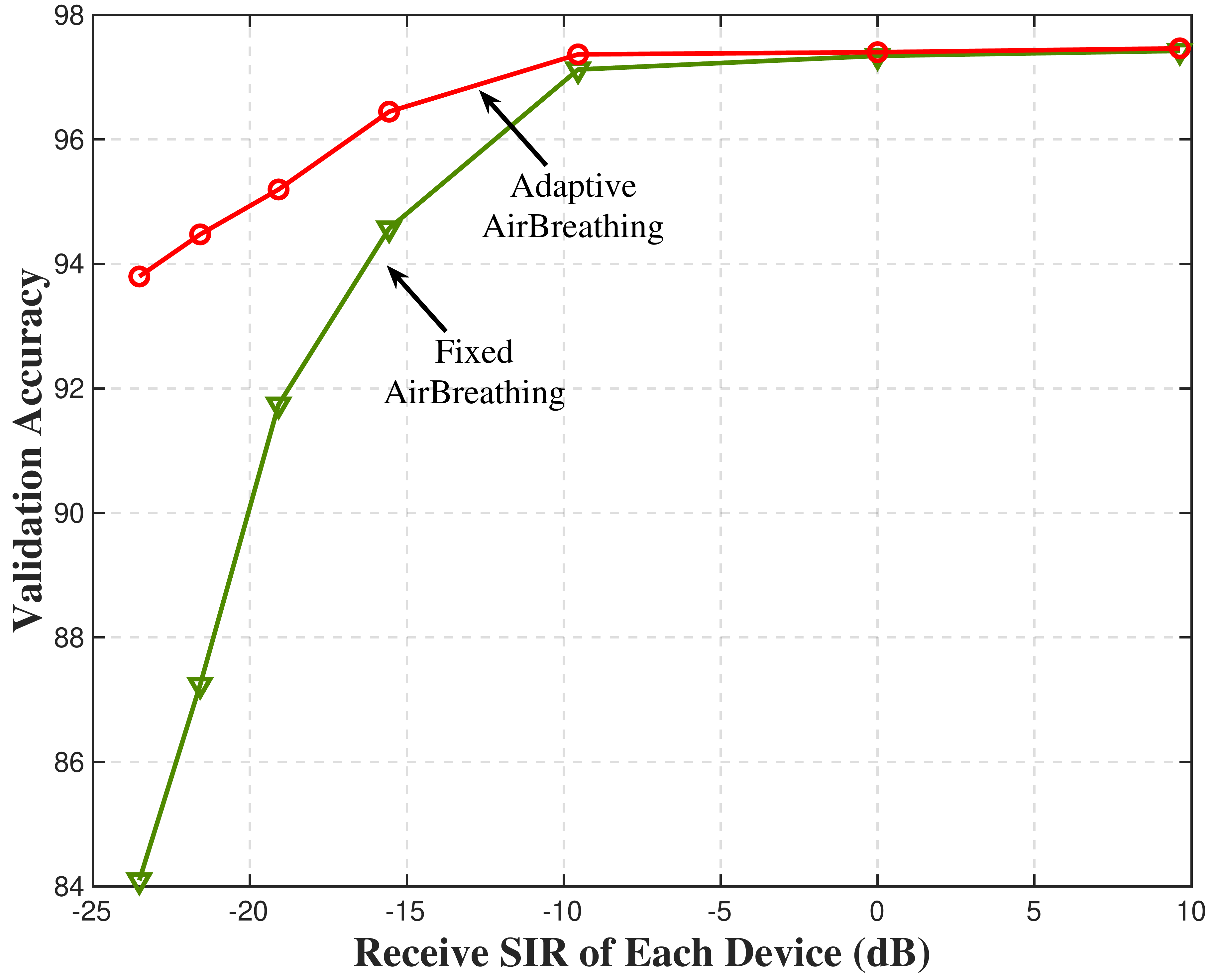}}
    \caption{The effects of (a) number of devices  and (b) receive SIR on the learning performance for given communication time of $7\times 10^6$ chips.}
    \label{fig:num_user-CNN-RC-SB}\vspace{-3mm}
\end{figure}

We study the effects of two key network parameters, namely the number of devices and the receive SIR per device, on the learning performance of AirBreathing FL for a given communication time of $7\times 10^6$ chips.
To this end, the curves of validation accuracy versus varying network parameters are plotted in Fig. \ref{fig:num_user-CNN-RC-SB}.

From Fig. \ref{fig:num_user-CNN-RC-SB} (a), one can see the continuous learning-performance improvement as the number of devices increases, as the AirComp's aggregation over devices suppresses interference by averaging.
Note that AirBreathing realizes interference suppression using a different mechanism of spread spectrum.
On the other hand, as expected, the AirBreathing FL with either fixed or adaptive breathing depth sees growing validation-accuracy improvement as the receive SIR per device (before aggregation) becomes larger.

\section{Conclusion}\label{Sec:conclusion}

In this work, we presented a spectrum-efficient method, called spectrum breathing. 
Leveraging the graceful degradation of learning performance due to pruning, the method exploits signal spectrum contraction via pruning to enable interference suppression via spread spectrum without requiring extra bandwidth.
The breathing depth that controls spectrum contraction level is optimized and adapted to both the states of gradient descent and channels to amplify the learning performance gain.

This work establishes a new principle of designing robust AirFL by integrating gradient pruning and interference suppression.
Beyond spread spectrum, this principle can be applied to other interference management techniques such as adaptive coding and modulation, MIMO beamforming, and cooperative transmission.
The current AirBreathing FL method can also be generalized to more complex systems such as multi-cell or distributed AirFL.
Practical issues such as synchronization errors and security warrant further investigation.

\appendix

\subsection{Proof of Lemma \ref{MSEG}}
\label{Proof of MSEG}
 In this section, the AirComp error of AirBreathing FL using random pruning is derived as below. For the expression brevity, the round index $n$ is omitted in the following equations.
First, the AirComp error, $\text{MSE}(n)$, is given as
\begin{equation}\small
    \begin{split}
&\text{MSE}(n)
=\mathbb{E}\left[\left\Vert \frac{1}{|\mathcal{K}|}\sum_{k\in\mathcal{K} }\mathbf{g}_k - \mathbf{y}^\prime \right\Vert^2\right]\\
=&\mathbb{E}_{\psi_n}\left[  \left\Vert \frac{1}{|\mathcal{K} |}\sum_{k\in\mathcal{K} }\mathbf{g}_k -R\left( \frac{1}{|\mathcal{K} |}\sum_{k\in\mathcal{K} }\mathbf{g}_k \right)\right\Vert^2 \right] +\mathbb{E}_{\widehat{\mathbf{z}}}\left[  \left\Vert R(\widehat{\mathbf{z}} ) \right\Vert^2 \right]\\
\overset{(a)}{=}& \mathbb{E}_{\psi_n}\left[  \left\Vert \frac{1}{|\mathcal{K} |}\sum_{k\in\mathcal{K} }\mathbf{g}_k -R\left( \frac{1}{|\mathcal{K} |}\sum_{k\in\mathcal{K} }\mathbf{g}_k \right)\right\Vert^2 \right]  +\frac{S_n P_I }{ G_n P_0 }\mathbb{E}\left[\frac{V^2}{|\mathcal{K}|^2}\right]\\
\overset{(b)}{=}&\left(1-\frac{S_n}{D} \right) \mathbb{E} \left[\left\Vert  \frac{1}{|\mathcal{K} |}\sum_{k\in\mathcal{K} }\mathbf{g}_k  \right\Vert^2\right] +\frac{S_n P_I }{ G_n P_0}\mathbb{E}\left[\frac{V^2}{|\mathcal{K}|^2}\right]\\
=&  \left(1-\gamma_n\right) \mathbb{E}[ \alpha^2(n)]  +  \frac{\gamma_n D  P_I}{ G_n P_0 }\mathbb{E}\left[\frac{V^2}{|\mathcal{K}|^2}\right],
    \end{split}
\end{equation}
where the expectation is taken over $\psi_n$ and $\widehat{\mathbf{z}}(n)$. 
$(a)$ is derived from the sum power of $S_n$ i.i.d zero mean Gaussian random variables. 
(b) is derived from the expectation of $\psi_n$ that is chosen at random from $\Omega_n$. That is, for a generic vector $\mathbf{x}\in \mathbb{R}^D$, the MSE between $\mathbf{x}$ and $R(\mathbf{x})$ over $\psi_n$ can be represented as \cite{RN100},
\begin{equation}
\label{Radom compression}
\begin{split}
     \mathbb{E}_{\psi_n}[\Vert\mathbf{x}-R(\mathbf{x})\Vert^2]
     =&\frac{1}{|\Omega_n|}\sum_{\psi_n\in \Omega_n}\sum_{d=1}^D [\mathbf{x}]_d^2\mathbb{I}\{d\notin \psi_n\}\\
     = & \sum_{d=1}^D [\mathbf{x}]_d^2 \sum_{\psi_n\in \Omega_n} \frac{\mathbb{I}\{d\notin \psi_n\}}{|\Omega_n|}\\
      =& \sum_{d=1}^D [\mathbf{x}]_d^2 \frac{{D-1 \choose S_n}}{{D \choose S_n}}\\
     =& (1-\frac{S_n}{D})\Vert \mathbf{x} \Vert^2=(1-\gamma_n)\Vert \mathbf{x} \Vert^2.
\end{split}
\end{equation}
$(b)$ trivially holds by replacing $\mathbf{x}$ of (\ref{Radom compression}) with $ \frac{1}{|\mathcal{K}|}\sum_{k\in\mathcal{K} }\mathbf{g}_k$.

\noindent The proof of Lemma \ref{MSEG} is completed.

\subsection{ Proof of Lemma \ref{MNEbound}}
\label{proof of MNEbound}
Consider round $n$, the gap between vanilla SGD and AirBreathing FL is upper bounded as 
\begin{equation}\small
\begin{split}
    &\mathbb{E}\left[\left\Vert \frac{1}{K}\sum_{k=1}^K\mathbf{g}_{k}(n)- \mathbf{y}^\prime(n) \right\Vert\right]\\
    =& \mathbb{E} \left[\left\Vert \frac{1}{K}\sum_{k=1}^K\mathbf{g}_{k}  -\frac{1}{|\mathcal{K} |}\sum_{k\in\mathcal{K} }\mathbf{g}_{k} +\frac{1}{|\mathcal{K} |}\sum_{k\in\mathcal{K} }\mathbf{g}_{k}  - \mathbf{y}^\prime \right\Vert\right]\\
    \leq &\mathbb{E} \left[\left\Vert \frac{1}{K}\sum_{k=1}^K\mathbf{g}_{k}  -\frac{1}{|\mathcal{K} |}\sum_{k\in\mathcal{K} }\mathbf{g}_{k} \right\Vert\right]+\mathbb{E} \left[\left\Vert\frac{1}{|\mathcal{K} |}\sum_{k\in\mathcal{K} }\mathbf{g}_{k}  - \mathbf{y}^\prime \right\Vert\right]\\
    \overset{(c)}{\leq} &\mathbb{E} \left[\left\Vert \frac{1}{K}\sum_{k=1}^K\mathbf{g}_{k}  -\frac{1}{|\mathcal{K} |}\sum_{k\in\mathcal{K} }\mathbf{g}_{k} \right\Vert\right]+\sqrt{\mathbb{E} \left[\left\Vert\frac{1}{|\mathcal{K} |}\sum_{k\in\mathcal{K} }\mathbf{g}_{k}  - \mathbf{y}^\prime \right\Vert^2\right]}\\
    = & \underbrace{\mathbb{E} \left[\left\Vert \frac{1}{K}\sum_{k=1}^K\mathbf{g}_{k}  -\frac{1}{|\mathcal{K} |}\sum_{k\in\mathcal{K} }\mathbf{g}_{k} \right\Vert\right]}_{\nu(n) }+\sqrt{\text{MSE}(n)},\\
\end{split}
\end{equation}
where $(c)$ comes from Jensen's inequality on concave function $\sqrt{\cdot}$; $\nu(n)$ is upper bounded as 
\begin{equation}
\begin{split}
     \nu(n)=&  \mathbb{E}\left[\left\Vert \frac{1}{K}\sum_{k=1}^K\mathbf{g}_{k} -\frac{1}{|\mathcal{K} |}\sum_{k\in\mathcal{K} }\mathbf{g}_{k} \right\Vert\right]\\
     = &\mathbb{E}\left[\left\Vert \frac{|\mathcal{K}|-K}{K|\mathcal{K}|}\sum_{k\in \mathcal{K}}\mathbf{g}_k+\frac{1}{K}\sum_{k\notin\mathcal{K}}\mathbf{g}_k\right\Vert\right]\\
       {\leq} & \sqrt{\mathbb{E}_{\mathbf{g}}\left[\left\Vert \frac{|\mathcal{K}|-K}{K|\mathcal{K}|}\sum_{k\in \mathcal{K}}\mathbf{g}_k+\frac{1}{K}\sum_{k\notin\mathcal{K}}\mathbf{g}_k\right\Vert^2\right]} \\
        \leq &\sqrt{\mathbb{E}_{|\mathcal{K}|}\left[\left(\frac{1}{|\mathcal{K}|}-\frac{1}{K}\right)\right]\sigma_g^2}
    \overset{(d)}{\leq} \sqrt{\frac{2-\xi_a}{K\xi_a}}\sigma_g,\\
\end{split}
\end{equation}
where $(d)$ is  due to that $|\mathcal{K}|$  subjects to binomial distribution $\mathcal{B}(K,\xi_a)$ such that the inequality below holds \cite{RN30}:
\begin{equation}
\label{expect of xi_a}
\begin{split}
    \mathbb{E}\left[\frac{1}{|\mathcal{K}|}\right]\leq \frac{2}{K \xi_a}, \quad
\mathbb{E}\left[\frac{1}{|\mathcal{K}|^2}\right]\leq \frac{6}{K^2 \xi_a^2}.
\end{split}
\end{equation}


\noindent The proof of Lemma \ref{MNEbound} is completed.

\subsection{Proof of Lemma \ref{U_n}}
\label{Proof of U_n}

We consider the process defined as follows
\begin{equation}
    U_n(\mathbf{w}(n),\dots,\mathbf{w}(0))\triangleq W_n(\mathbf{w}(n),\dots,\mathbf{w}(0))-\eta H\sum_{i=0}^{n-1}u(i),
\end{equation}
where $u(i)$ is given in Lemma \ref{MNEbound}, and global model is updated using (\ref{rece_signal}).
When the algorithm has not entered the success region at round $n$, we have $\forall n\geq0$,
\begin{equation}
    \begin{split}
          &U_{n+1}(\mathbf{w}(n+1),\dots,\mathbf{w}(0))\\
          =& W_{n+1}(\mathbf{w}(n)-\eta \mathbf{y}^\prime(n),\mathbf{w}(n),\dots,\mathbf{w}(0))-\eta H\sum_{i=0}^{n}u(i)\\
          \overset{(\ref{H-Lip})}{\leq}& W_{n+1}\left(\mathbf{w}(n)-\eta\frac{1}{K}\sum_{k=1}^K\mathbf{g}_k(n),\mathbf{w}(n),\dots,\mathbf{w}(0)\right)\\ 
          &+\eta H \left\Vert \frac{1}{K}\sum_{k=1}^K\mathbf{ g}_k(n)-\mathbf{y}^\prime(n)\right\Vert -\eta H\sum_{i=0}^{n}u(i),
    \end{split}
\end{equation}
where the scaling up results from the $H$-Lipschitz smooth in the first coordinate given in (\ref{H-Lip}).

Then we take the expectation for both sides of the inequality and use the supermartingle property of $W_n$. The expectation of $U_{n+1}$ is bounded by
\begin{equation}
    \begin{split}
          &\mathbb{E}[U_{n+1}(\mathbf{w}(n+1),\dots,\mathbf{w}(0))]\\
          \leq &\mathbb{E}\left[W_{n+1}\left(\mathbf{w}(n)-\eta\frac{1}{K}\sum_{k=1}^K\mathbf{ g}_k(n),\mathbf{w}(n),\dots,\mathbf{w}(0)\right)\right]\\ 
          &+\eta H \mathbb{E}\left[ \left\Vert \frac{1}{K}\sum_{k=1}^K\mathbf{g}_k(n)- \mathbf{y}^\prime(n) \right\Vert \right ]-\eta H\sum_{i=0}^{n}\mathbb{E}[u(i)]\\
          \leq& W_n(\mathbf{w}(n),\dots,\mathbf{w}(0))+\eta H u(n)-\eta H\sum_{i=0}^{n}u(i)\\
          =&U_n(\mathbf{w}(n),\dots,\mathbf{w}(0)).
    \end{split}
\end{equation}
The inequality above still holds in the case when the algorithm has succeeded at round $n$. 
Thus, $U_n$ is a supermartingale process for AirBreathing FL. 

\noindent Proof of Lemma \ref{U_n} is completed.

\subsection{Proof of Theorem \ref{The_1}}
\label{proof of The_1}
We denote the failure to enter the success region by $N$ as $F_N$, otherwise, the success as $\neg F_N$. Consider the same model initialization $\mathbf{w}(0)$ for $U_n$ and $W_n$, we  have
\begin{equation}
    \begin{split}
        \mathbb{E}[W_0]&=\mathbb{E}[U_0]
        \geq\mathbb{E}[U_n]\\
        &=\mathbb{E}[U_n|F_N]\text{Pr}\{F_N\}+\mathbb{E}[U_n|\neg F_N] \text{Pr}\{\neg F_N\}\\
        &\geq\mathbb{E}[U_n|F_N]\text{Pr}\{F_N\}\\
        &=\left(\mathbb{E}[W_N|F_N]-\eta H \sum_{n=0}^{N-1}u(n)\right)\text{Pr}\{F_N\}\\
        &\geq \left(N-\eta H \sum_{n=0}^{N-1}u(n)\right)\text{Pr}\{F_N\}.
    \end{split}
\end{equation}
Hence, we obtain 
\begin{equation}
    \begin{split}
        \text{Pr}\{F_N\}\leq\frac{ \mathbb{E}[W_0]}{N-\eta H \sum_{n=0}^{N-1}u(n)},
    \end{split}
    \label{Prob F_N}
\end{equation}
where $\mathbb{E}[W_0]$ can be obtained by setting $n=0$ in (\ref{W_t expression}) and taking expectation.

\noindent Proof of theorem \ref{The_1} is completed.

\subsection{Proof of Lemma \ref{alpha+V}}
\label{Proof of alpha+V}
The round index is omitted for expression brevity. By taking the expectation over the gradient and number of active devices, the upper bound of $\mathbb{E}[\alpha^2(n)]$  is given as
\begin{equation}
    \begin{split}
    \mathbb{E}[\alpha^2(n)]
     =&\mathbb{E}\left[\left\Vert \frac{1}{|\mathcal{K}|}\sum_{k\in \mathcal{K}}\mathbf{g}_k \right\Vert^2\right]
     \leq  \mathbb{E}\left[ \Vert \mathbf{g}\Vert^2+\frac{\sigma_g^2}{|\mathcal{K}|}\right]\\
     \overset{(\ref{expect of xi_a})}{\leq}& \Vert \mathbf{g}\Vert^2+\frac{2\sigma_g^2}{K\xi_a} \overset{(e)}{\leq}\Vert \mathbf{g}\Vert^2+\sigma_g^2,
    \end{split}
\end{equation}
where $(e)$ trivially holds when no less than 2 devices are expected to be active at each round such that $K\xi_a\geq2$.
By taking the expectation over selected element set $\mathcal{\psi}_n$, $ \mathbb{E}\left[V^2(n)\right] $ is upper bounded as below:
\begin{equation}
\label{eq:above}
    \begin{split}
   \mathbb{E}\left[V^2(n)\right]
    &= \mathbb{E}\left[ \frac{1}{|\mathcal{K}|}\sum_{k\in\mathcal{K}}V_k^2 \right] \\
         &=\mathbb{E}\left[ \frac{1}{S_n}\sum_{s=1}^{S_n} [\mathbf{g}^{\text{co}}]^2_s-M^2_k\right]\\
         &\leq\mathbb{E}\left[ \frac{1}{S_n}\sum_{s=1}^{S_n} [\mathbf{g}^{\text{co}}]^2_s\right]\\
         &= \frac{1}{D}\sum_{d=1}^D [\mathbf{g}_k]_{d}^2\frac{D}{S_n}
         \sum_{\psi_n\in \Omega_{n}}\frac{ \mathbb{I}\{d\in \psi_n\}}{|\Omega_{n}|}\\
         &= \frac{1}{D}\Vert \mathbf{g}_k \Vert^2 \frac{D}{S_n}\frac{{D-1 \choose S_n-1}}{{D \choose S_n}}
         = \frac{1}{D}\Vert \mathbf{g}_k \Vert^2\\
         &\leq \frac{1}{D}(\Vert \mathbf{g}\Vert^2+\sigma_g^2 ),
    \end{split}
\end{equation}
where $M_k=\frac{1}{S_n}\sum_{s=1}^{S_n} [\mathbf{g}_k^{\text{co}}]_s$ is the mean of local sparse gradient. Building on the analysis above, $ \mathbb{E}[\alpha^2(n)]\leq D\Gamma(n)$  and $\mathbb{E}[V^2(n)]\leq \Gamma(n)$  holds if the upper bound of GSI is chosen as $\Gamma(n) \geq \frac{1}{D}(\Vert \mathbf{g}\Vert^2+\sigma_g^2 ) $. Last, the upper bound for CSI  can be simply obtained by (\ref{expect of xi_a}).

\noindent Proof of Lemma \ref{alpha+V} is completed.

\bibliography{main}
\bibliographystyle{IEEEtran}

\end{document}